\documentclass{article}

\usepackage{microtype}
\usepackage{graphicx}
\usepackage{caption}
\usepackage{wrapfig}
\usepackage{subcaption}
\usepackage{booktabs}
\usepackage{hyperref}
\usepackage{tikz}
\usetikzlibrary{positioning,arrows.meta,shapes.geometric,fit,backgrounds,calc,decorations.pathreplacing}
\usepackage{xcolor}
\usepackage{xspace}
\usepackage[most]{tcolorbox}

\usepackage[preprint]{icml2026}

\usepackage{amsmath}
\usepackage{amssymb}
\usepackage{mathtools}
\usepackage{amsthm}
\usepackage[capitalize,noabbrev]{cleveref}
\usepackage{xfp}
\usepackage{makecell}

\theoremstyle{plain}

\theoremstyle{definition}

\theoremstyle{remark}

\definecolor{imgblue}{RGB}{66,133,244}
\definecolor{blockgreen}{RGB}{52,168,83}
\definecolor{maskorange}{RGB}{251,188,5}
\definecolor{tokengray}{RGB}{200,200,200}
\definecolor{eosred}{RGB}{234,67,53}
\definecolor{diffpurple}{RGB}{156,39,176}
\usepackage{pgfplots}
\pgfplotsset{compat=1.18}
\usepackage{multirow}
 
\makeatletter
\DeclareRobustCommand\onedot{\futurelet\@let@token\@onedot}
\def\@onedot{\ifx\@let@token.\else.\null\fi\xspace}

 \def\vs{\emph{vs}\onedot}

\makeatother

\icmltitlerunning{DODO: Discrete OCR Diffusion Models}

\newcommand{\QualitativeRow}[3]{%
    \begin{subfigure}[t]{0.32\linewidth}
        \centering
        \includegraphics[width=\linewidth, height=6cm, keepaspectratio]{#1}
    \end{subfigure}
    \hfill
    \begin{subfigure}[t]{0.32\linewidth}
        \centering
        \includegraphics[width=\linewidth, height=6cm, keepaspectratio]{#2}
    \end{subfigure}
    \hfill
    \begin{subfigure}[t]{0.32\linewidth}
        \centering
        \includegraphics[width=\linewidth, height=6cm, keepaspectratio]{#3}
    \end{subfigure}
    \\[1em]
}

\begin{document}

\twocolumn[
    \icmltitle{
DODO: Discrete OCR Diffusion Models
}
    \icmlsetsymbol{equal}{*}
    
    \begin{icmlauthorlist}
    \icmlauthor{Sean Man$^*$}{1}
    \icmlauthor{Gilad Deutch$^*$}{2}
    \icmlauthor{Roy Ganz}{2}
    \icmlauthor{Roi Ronen}{2}
    \icmlauthor{Shahar Tsiper}{2}
    \icmlauthor{Shai Mazor}{2}
    \icmlauthor{Niv Nayman}{2}
    \end{icmlauthorlist}
    
    \icmlaffiliation{1}{Technion - Israel Institute of Technology, Haifa, Israel.}
    \icmlaffiliation{2}{Amazon Web Services}
    
    \icmlcorrespondingauthor{Niv Nayman}{nivay@amazon.com}
    
    \icmlkeywords{Diffusion Models, OCR, Vision-Language Models, Document Understanding}
    
    \vskip 0.3in
]
\def\thefootnote{*}\footnotetext{These authors contributed equally to this work. Work conducted during an internship at Amazon}
\printAffiliationsAndNotice{}

\begin{abstract}
Optical Character Recognition (OCR) is a fundamental task for digitizing information, serving as a critical bridge between visual data and textual understanding.
While modern Vision-Language Models (VLM) have achieved high accuracy in this domain, they predominantly rely on autoregressive decoding, which becomes computationally expensive and slow for long documents as it requires a sequential forward pass for every generated token.
We identify a key opportunity to overcome this bottleneck: unlike open-ended generation, OCR is a highly deterministic task where the visual input strictly dictates a unique output sequence, theoretically enabling efficient, parallel decoding via diffusion models.
However, we show that existing masked diffusion models fail to harness this potential; those introduce structural instabilities that are benign in flexible tasks, like captioning, but catastrophic for the rigid, exact-match requirements of OCR.
To bridge this gap, we introduce \textbf{DODO}, the first VLM to utilize block discrete diffusion and unlock its speedup potential for OCR.  
By decomposing generation into blocks, DODO mitigates the synchronization errors of global diffusion.
Empirically, our method achieves near state-of-the-art accuracy while enabling up to $5\times$ faster inference compared to autoregressive baselines.
\end{abstract}

\section{Introduction}
\label{sec:intro}

\begin{figure}[ht]
    \centering
    \includegraphics[width=0.9\linewidth]{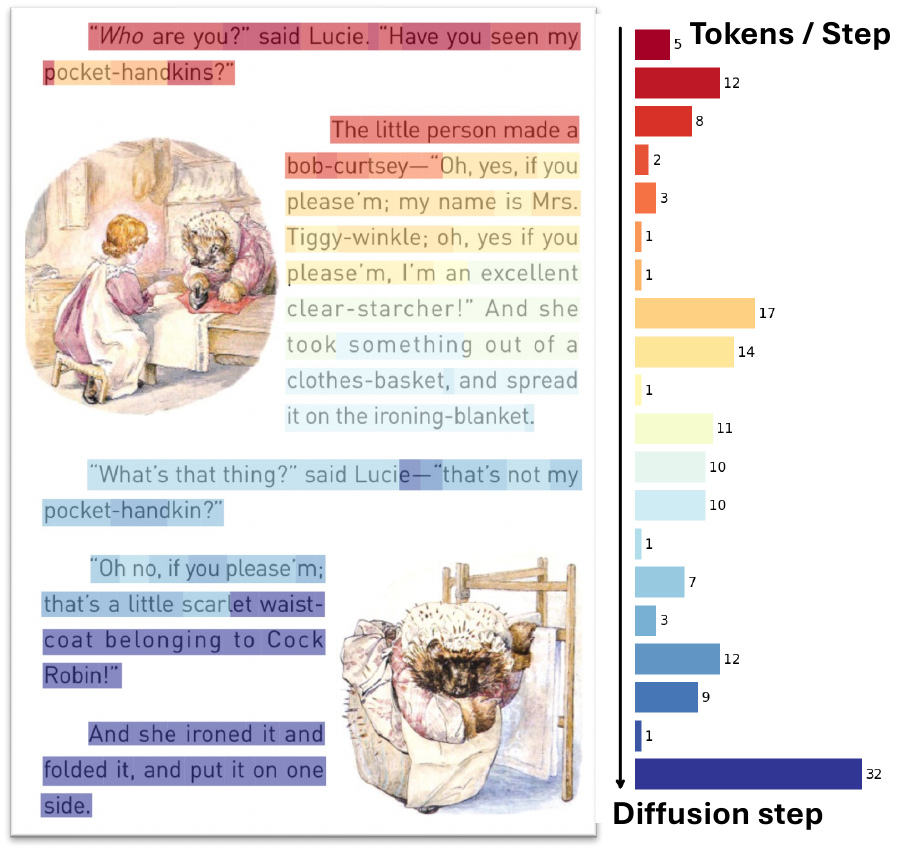}
    \caption{
\textbf{DODO: High-throughput parallel generation.} 
Unlike autoregressive models constrained to a strict left-to-right sequence, DODO generates text across the entire canvas simultaneously (with same color) based on visual confidence.
In this example, it resolves $148$ tokens in just $20$ forward passes ($\approx 7$ tokens/step on average).
Simultaneously generated tokens are adaptively distributed across steps.}
    \label{fig:token_order}
\end{figure}

Optical character recognition (OCR) is a core component of modern document understanding systems, enabling the extraction of structured text from images such as scanned documents, forms, and natural scenes. 
Vision--language models are increasingly used for large-scale document parsing and multimodal reasoning~\cite{NEURIPS2022_960a172b,pmlr-v162-li22n,li2023blip2bootstrappinglanguageimagepretraining,Ganz_2023_ICCV,NEURIPS2023_9a6a435e,wu2024deepseekvl2mixtureofexpertsvisionlanguagemodels,li2024llavaonevisioneasyvisualtask,ganz2024questionawarevisiontransformer,chen2025expandingperformanceboundariesopensource,bai2025qwen25vltechnicalreport,bai2025qwen3vltechnicalreport,liu2025nvilaefficientfrontiervisual}.
However, the high computational cost and latency of these architectures have re-established OCR transcription as a critical bottleneck where both accuracy and inference efficiency are essential~\cite{blecher2023nougatneuralopticalunderstanding,wei2024generalocrtheoryocr20,abramovich2024visfocus,nacson2025docvlm, wei2025deepseekocrcontextsopticalcompression}.

Crucially, OCR differs fundamentally from semantically flexible tasks like image captioning~\cite{sidorov2020textcapsdatasetimagecaptioning,chen2023sharegpt4vimprovinglargemultimodal,lin2014microsoft} or visual question answering (VQA)~\cite{antol2015vqa,singh2019vqamodelsread,mathew2021docvqa,yue2024mmmumassivemultidisciplinemultimodal} as it is semantically rigid.
Conditioned on the image, the posterior distribution is effectively unimodal, meaning the visual input strictly dictates a single valid sequence.
This determinism exposes a critical inefficiency in standard Autoregressive (AR) models: they generate text sequentially, creating a significant latency bottleneck for long document sequences.
Conversely, this characteristic makes OCR uniquely suited for Masked Diffusion Models (MDMs)~\cite{sahoo2024simpleeffectivemaskeddiffusion}.
Because the output allows for little ambiguity, OCR satisfies the MDM assumption of \textit{conditional independence}—the premise that tokens can be predicted independently given the input~\cite{azangulov2025parallelsamplingmaskeddiffusion}.
Theoretically, this allows the model to resolve large spans of text simultaneously, similar to how traditional OCR pipelines~\cite{wang2021object, ronen2022glass, aberdam2023clipter} recognize isolated regions in parallel without the risk of incoherence.

However, realizing this potential in practice reveals a structural paradox: the same rigidity that enables parallelization also makes OCR particularly sensitive to the instabilities of global decoding. While standard MDMs~\cite{dimple,lavida,lladav} can generate tokens in parallel, they introduce non-causal structural uncertainties, specifically regarding sequence length and absolute positional alignment.
In flexible tasks like captioning, such errors are recoverable: the model can navigate a wide space of valid outputs to resolve a misalignment dynamically.
OCR allows no such flexibility.
Because the target is a single, immutable sequence, structural errors become irrecoverable; the model cannot ``rewrite'' the text to compensate for incorrect length estimates or token placement.
Consequently, these rigidities force the model to either truncate valid text or hallucinate padding, leading to fractured, colliding outputs that fundamentally undermine the efficacy of standard masked diffusion for transcription.

To resolve this paradox, we propose DODO (\underline{D}iscrete \underline{O}CR \underline{D}iffusion M\underline{o}dels), the first Vision--Language Model to adapt block discrete diffusion for document transcription. Unlike standard global diffusion, DODO decomposes the monolithic generation task into a sequence of causally anchored blocks. This structural change directly addresses the rigidities of OCR: by bounding the inference horizon and conditioning on a committed prefix, we eliminate the risk of long-range alignment drift and enable dynamic length adaptation without requiring a perfect global estimate. 
Crucially, we leverage the high-confidence nature of OCR combined with block-causal attention and KV-caching~\cite{li2024llavaonevisioneasyvisualtask} to achieve both causal consistency and fast inference.
Moreover, unlike autoregressive models, the diffusion framework uniquely enables bidirectional attention across blocks, which we show pushes the utilizable block size from $32$ to $256$ tokens, further improving accuracy.

Empirically, DODO achieves transcription accuracy competitive with state-of-the-art autoregressive models while outperforming the equivalent autoregressive baseline in throughput.
These results validate our hypothesis: OCR is indeed a regime where the conditional independence assumption holds, but it requires the structural safety rails of block diffusion to be realized in practice.
Appropriately structured, DODO recovers the correctness of AR models while unlocking the efficiency benefits that motivate diffusion-based decoding in the first place.

Our contributions are summarized as follows:
\begin{itemize}
\item We identify a structural incompatibility between standard masked diffusion and the rigid requirements of OCR, explaining why positional and length errors that are benign in flexible tasks prove catastrophic for OCR.
\item We introduce \textbf{DODO}, the first VLM to utilize block discrete diffusion. By decomposing generation into sequentially conditioned blocks, DODO enforces local alignment and enables dynamic length adaptation, resolving the rigidities of global diffusion.
\item We demonstrate that DODO matches the accuracy of state-of-the-art autoregressive baselines while enabling up to $5\times$ faster inference, validating the potential of parallel decoding for dense text recognition.
\end{itemize}

\section{Related Work}
\label{sec:related}

\begin{figure*}[t]
    \centering
    \includegraphics[width=\linewidth]{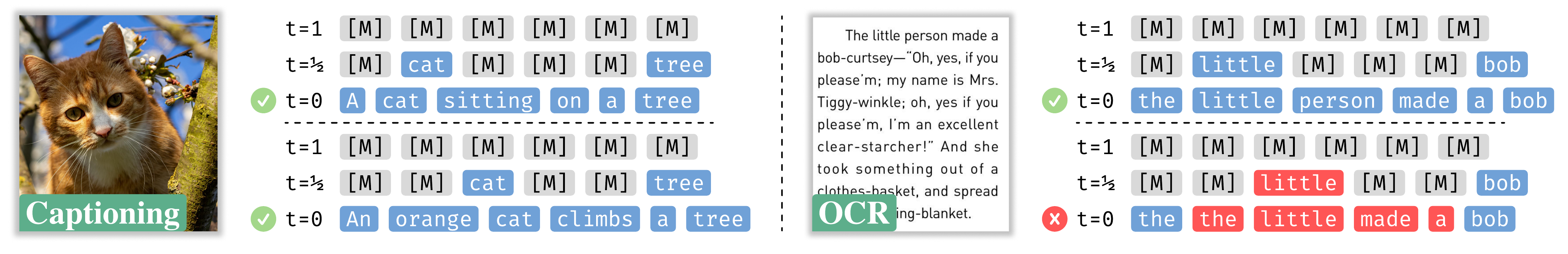}
    \caption{
\textbf{Semantically flexible \vs semantically rigid vision–language tasks.}
\textit{Left:} Image captioning admits multiple, semantically equivalent descriptions of the same image. Different decoding trajectories can converge to distinct but equally valid captions, and lexical or structural variations are naturally absorbed.
\textit{Right:} OCR requires a single, exact transcription determined by the image. Even minimal local deviations, such as an incorrect token choice or boundary, render the output incorrect.
As a result, conditioned on the image, OCR exhibits extremely low output variability, which makes it a natural candidate for parallel decoding, but also a demanding setting in which errors cannot be compensated by alternative phrasings or later corrections.
}
    \label{fig:ocr_captioning}
\end{figure*}

\paragraph{Specialized OCR and Document Understanding.}
Modern OCR systems leverage vision-language models for end-to-end document understanding. MonkeyOCR~\citep{monkeyocr} introduces a multi-stage pipeline with detection, recognition, and reading order prediction. MinerU~\citep{wang2024mineru} and commercial systems like dots.ocr~\cite{li2025dotsocrmultilingualdocumentlayout}, DeepSeek-OCR~\cite{wei2025deepseek}, Mistral-OCR~\cite{mistral2025ocr} achieve strong performance through careful engineering and large-scale training. These methods universally employ autoregressive decoding. This is the first successful attempt to achieve competitive OCR performance by MDMs.

\paragraph{Discrete Diffusion Models.}
Discrete diffusion models learn to reverse a corruption process over discrete tokens. D3PM~\citep{d3pm} introduced structured transition matrices, while MDLM~\citep{mdlm} simplified training through masked diffusion with a tighter evidence lower bounds (ELBO). Recent work has scaled these models to language modeling~\citep{dream}, though a gap remains compared to autoregressive models on perplexity benchmarks. This work narrows down the performance gap of MDMs with their plain autoregressive counterparts for the studied task.

\paragraph{Block Diffusion.}
Block Diffusion (BD3-LM)~\citep{blockdiffusion} bridges autoregressive and diffusion models by generating blocks of tokens autoregressively during inference only, with each block decoded via masked diffusion. This enables KV-caching across blocks while maintaining parallel decoding within blocks. Prior work uses small block sizes (4--32 tokens)~\cite{fastdllm} to minimize the performance gap with autoregressive models on language modeling. \cite{wu2025fast} further aligns the attention masks to be block causal during training. We implement this approach for VLMs.

\paragraph{Multimodal Diffusion Models.}
Dimple~\citep{dimple} extends discrete diffusion to vision-language tasks, training on the LLaVA recipe. However, it shows limited gains over autoregressive baselines and does not evaluate on OCR benchmarks. LaViDa~\citep{lavida} and LLaDA-V~\citep{lladav} explore diffusion for VQA but struggle with OCR tasks requiring precise text reproduction. This work is the first to successfully apply discrete diffusion for OCR.

\section{Preliminaries}
\label{sec:prelim}

\paragraph{Notation.}
Let $\mathcal{V}$ be a vocabulary of size $V$, and let $\mathtt{[M]}\notin\mathcal{V}$ denote a dedicated  \texttt{MASK} token.
We write $\tilde{\mathcal{V}}=\mathcal{V}\cup\{\mathtt{[M]}\}$, and represent tokens either as categorical indices $v \in \tilde{\mathcal{V}}$ or as one-hot vectors $\mathbf{e}_v \in \{0,1\}^{|\tilde{\mathcal{V}}|}$.
We use $\mathrm{Cat}(\cdot;\pi)$ for a categorical distribution with probabilities $\pi$.

\subsection{OCR as Conditional Sequence Modeling}
\label{sec:prelim_ocr}


We formulate OCR as a conditional generation task where the goal is to map a document image $I$ to a target sequence of discrete tokens $x^{1:L}$.
This target is defined as $x^{1:L} = \tau(s(I))$, where $s(\cdot)$ represents a fixed serialization scheme (e.g., plain text, \LaTeX{}, HTML) and $\tau(\cdot)$ is a tokenizer that maps the resulting string to vocabulary indices.
A vision--language model (VLM) estimates the conditional distribution $p_\theta(x^{1:L}\mid I, c)$ given the image $I$ and optional text context $c$.
Autoregressive (AR) VLM decoding admits standard left-to-right factorization
\begin{equation}
    \log p_\theta(x^{1:L}\mid I,c)=\sum_{\ell=1}^L \log p_\theta(x^\ell \mid x^{<\ell}, I,c),
    \label{eq:elbo}
\end{equation}
with $x^{<l}$ the prefix tokens at step $l$ out of $L$ sequential steps.


\subsection{Masked Diffusion Models (MDMs)}
\label{sec:prelim_mdm}

\paragraph{Forward (Masking) Process.}
MDMs define a coordinate-independent corruption process that replaces tokens with $\mathtt{[M]}$ according to a noise level $t\in[0,1]$. Writing $x_0$ for clean data and $x_t$ for its noisy version, a common conditional probability distribution for the token masking is
\begin{equation}
    q_{t|0}(x_t\mid x_0)=\prod_{i=1}^L \mathrm{Cat}\!\Big(x_t^i;\ \alpha_t\,\mathbf{e}_{x_0^i} + (1-\alpha_t)\,\mathbf{e}_{\mathtt{[M]}}\Big),
\end{equation}
where $\alpha_t$ is strictly decreasing with $\alpha_0\!=\!1$ and $\alpha_1\!=\!0$.

\paragraph{Training.}
MDMs train a denoiser to predict the original input tokens from partially masked ones. In continuous time, an ELBO-derived objective can be written \cite{mdlm, simpledlm} as a weighted masked cross-entropy
\begin{equation}
    \mathcal{L} = \mathbb{E}_{t, x_0, x_t|x_0} \left[ \frac{\alpha'_t}{1-\alpha_t} \sum_{i: x_t^i = \texttt{[M]}} -\log p_\theta(x_0^i | x_t, t) \right],
\end{equation}
where $\alpha'_t=\frac{d\alpha_t}{dt}$, and $p_\theta(\cdot\mid x_t,t)$ is frequently implemented without an explicit time embedding since $x_t$ reveals $t$ through its mask rate.

\paragraph{Sampling.}
MDM sampling starts from the fully masked sequence $x_1=(\mathtt{[M]},\dots,\mathtt{[M]})$ and iterates noise levels $1=t_K>\cdots>t_0=0$.
Given an estimate of the marginal distribution of each token from the denoiser, a common and convenient decomposition~\cite{mdlm} of a single reverse step $t_{k+1} \to t_k$ proceeds by \emph{first} choosing which masked positions to reveal via a selection rule (e.g., randomly~\cite{mdlm}, top-k~\cite{zheng2023reparameterized}, confidence-thresholding~\cite{dimple}, or deterministically~\cite{dus}), and \emph{second}, sampling token values for those positions from the denoiser's predicted distribution $x_{t_k}^i\sim p_\theta(x_{t_k}^i\mid x_{t_{k+1}})$.

This decomposition has two consequences described next.

\begin{figure}[t]
    \centering
    \includegraphics[width=\linewidth]{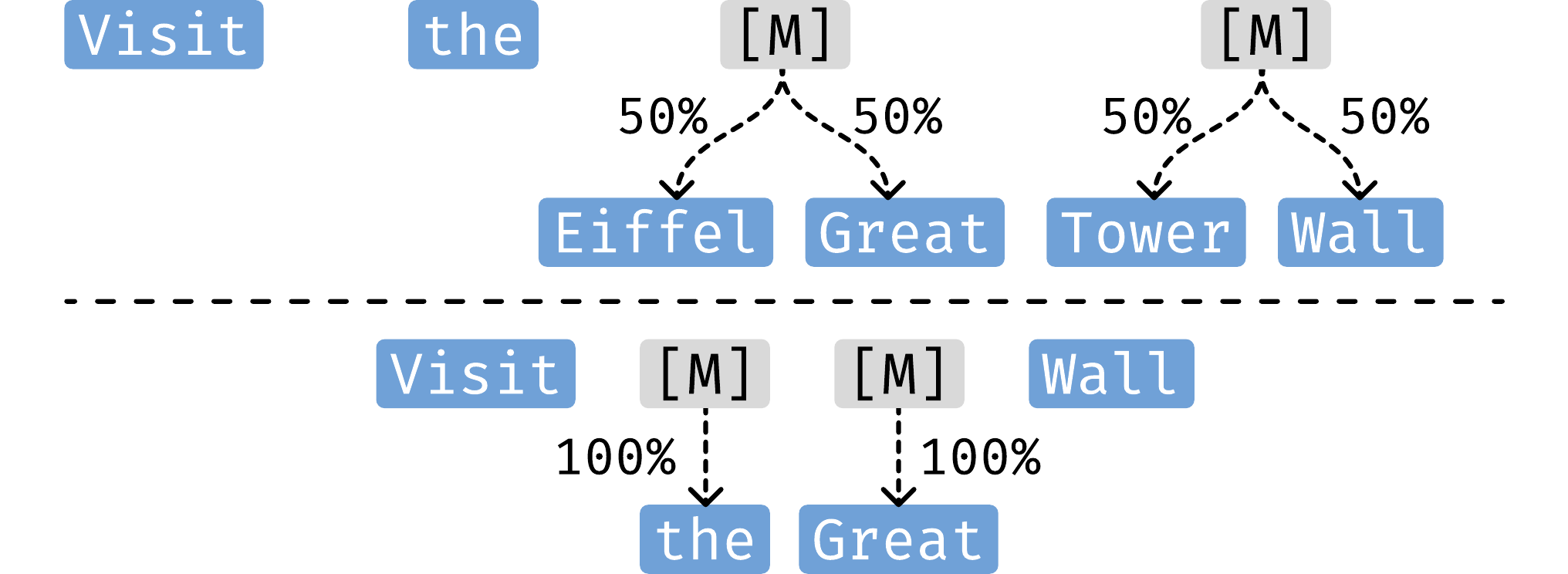}
    \caption{
    \textbf{Conditional independence assumption.}
    \emph{(Top)} In open-ended tasks, ambiguity between valid options risks sampling incoherent mixtures.
    \emph{(Bottom)} In OCR, the strong visual signal resolves this ambiguity, enabling conflict-free parallel decoding.
    \label{fig:conditional_independence}
}
\end{figure}

\paragraph{Conditional Independence Assumption.}
At each sampling step, the decoded tokens are sampled independently from one another.
This may lead to incorrect results if the decoded tokens in fact are conditionally dependent given the context, as illustrated in \cref{fig:conditional_independence}.
On the other hand, when this assumption holds, there is a large parallelization potential to leverage.


\paragraph{Carry-Over Unmasking.} 
\label{para:carry-over}
 We follow the common practice \cite{fastdllm, wu2025fast, lladav, lavida, dimple} and only allow the sampling of masked tokens at each step.
While this formulation is shown by \citep{mdlm} to be the key for the derivation of the discrete ELBO equivalent loss in \cref{eq:elbo}, this means one cannot revise previously decoded tokens. 
While this has a lesser effect on generative tasks where multiple responses are acceptable, it might be detrimental for tasks where only one response is correct, as illustrated in \cref{fig:ocr_captioning}.

\section{Method}
\label{sec:ocr_mdm}

We analyze OCR through the lens of MDM training and sampling (\cref{sec:prelim_mdm}), focusing on the two sampling assumptions made explicit there: \emph{conditional independence} of tokens sampled within a step, and \emph{carry-over unmasking}, where revealed tokens are not revised.

We first argue that OCR is especially compatible with the conditional-independence assumption.
We then show why this potential is difficult to realize with vanilla MDM inference, as early mistakes persist and this requires caution when decoding many tokens in parallel.
Finally, we argue that block discrete diffusion mitigates these failure modes, while retaining high parallelism and enjoying KV-Caching.

\subsection{Parallel Decoding Potential}
\label{sec:ocr_parallelism}

OCR typically yields long sequences, making standard autoregressive decoding a significant latency bottleneck, as it requires $L$ sequential forward passes to decode $L$ tokens.
Masked diffusion models offer a compelling alternative by enabling parallel token generation; however, their effectiveness hinges on the validity of the \emph{conditional independence assumption}.

We argue that OCR is uniquely suited for this parallel paradigm.
Unlike semantically flexible vision-language tasks, the posterior distribution $p(x^{1:L}|I,c)$ for document transcription is highly peaked, often approaching a Dirac delta function around a single ground-truth sequence.
In this low-entropy regime, the strong conditioning on the visual input effectively decouples token predictions, allowing the joint probability of masked tokens to be factorized:
\begin{equation}
    p(x^{1:L}_{t_k} \mid x^{1:L}_{t_{k+1}}, I, c) \approx \prod_{\ell=1}^L p(x^\ell_{t_k} \mid x^{1:L}_{t_{k+1}}, I, c).
\end{equation}
This independence allows the decoding of large spans of tokens simultaneously, as shown in \cref{fig:token_order}.

\subsection{Brittleness of Parallel Decoding in OCR}
\label{sec:parallel_rigidity}

Standard masked diffusion models operate on a fixed-length canvas and rely on \emph{carry-over unmasking} (\cref{para:carry-over}), treating tokens revealed in early steps as immutable context.
While this constraint is manageable for semantically flexible tasks like captioning, where the model can paraphrase content or alter semantics to fit the available space, it presents a fundamental challenge for OCR.
Document transcription is a rigid, zero-tolerance task: the ground-truth text consists of a specific set of characters in an immutable order.
This lack of flexibility exposes two critical failure modes for parallel decoding:

\paragraph{Length Mismatch.}
Because the true sequence length is unknown at inference time, the decoding canvas size is effectively an estimate.
In generative tasks like captioning, this is rarely fatal; the model can simply generate a valid shorter or longer description to match the canvas.
In OCR, however, this mismatch creates a structural vulnerability.
If the initial $L$ is incorrect, or the model predicts an end-of-sequence \texttt{[EOS]} token prematurely, the model is forced to either truncate valid text (if the effective length is too short) or hallucinate (if too long) to satisfy the imposed constraint.

\paragraph{Positional Anchoring.}
Even with a valid length estimate, parallel decoding binds content to absolute positional indices in a non-causal manner.
This introduces a critical synchronization risk: the model may predict a segment at an incorrect offset, such as placing a table header $50$ tokens too early or too late.
Because \emph{carry-over unmasking} prevents the revision of revealed tokens, this error is locked in place.
Unlike autoregressive decoding, which inherently align content to its history without looking ahead, diffusion parallel decoding is bound to errors made ahead.
Consequently, the subsequent text cannot shift to accommodate the offset.
Due to the unimodal nature of OCR, the model cannot simply paraphrase or ``tweak'' the surrounding text to bridge this misalignment, leading to fractured outputs where disjoint segments collide -- a fundamental challenge that limits the efficacy of purely parallel OCR.

\subsection{Block Diffusion as a Structural Remedy}
\label{sec:ocr_block_tease}

\begin{figure}
    \centering
    \includegraphics[width=1\linewidth]{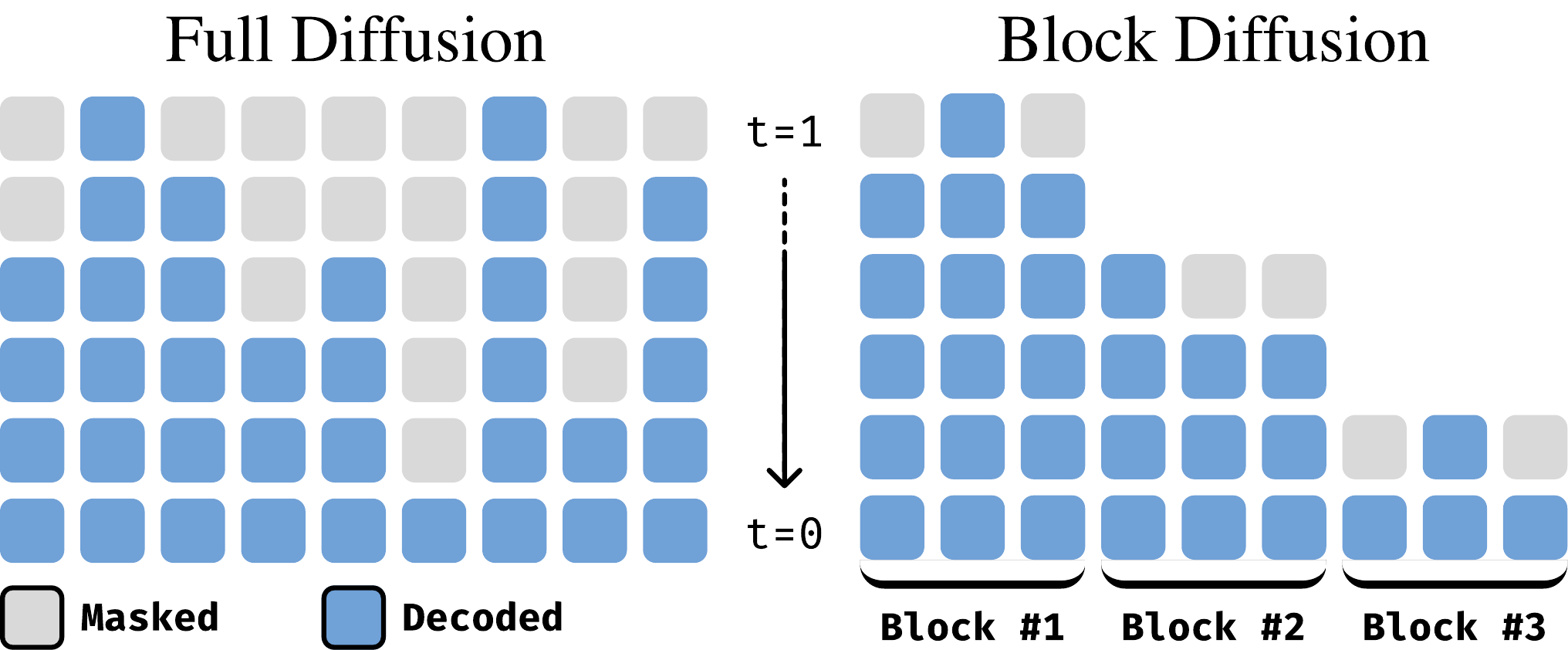}
    \caption{
    \textbf{Full \vs block diffusion}. In standard full diffusion (\emph{left}), MDM sampling is applied globally to the entire sequence. In contrast, block diffusion (\emph{right}) restricts parallel sampling to discrete windows, processing blocks sequentially from left to right.
    }
    \label{fig:block_full}
\end{figure}
\vspace{-3mm}

Block discrete diffusion can mitigate these failure modes by replacing a single length-$L$ denoising problem with a sequence of bounded-span problems, conditioned of a prefix composed on previous blocks decoded sequentially.

Block discrete diffusion models~\cite{blockdiffusion} combine AR structure at a coarse granularity with diffusion within blocks. Partition the sequence into $B$ contiguous blocks of length $L'$ (so $L=BL'$) and write $x^{(b)}$ for block $b$ and $x^{(<b)}$ for the prefix blocks. The model factorizes as
\begin{equation}
    p_\theta(x^{1:L}| I, c) = \prod_{b=1}^B p_\theta\!\big(x^{(b)} \mid x^{(<b)}, I, c\big),
\end{equation}
where each conditional distribution $p_\theta(x^{(b)} \mid x^{(<b)}, I, c)$ is implemented via a masked diffusion sampling over the $L'$ tokens of the block, conditioned on prefix states.
This formulation narrows the performance gap between MDMs and AR models and enables significant computational efficiency by allowing the KV-cache of committed prefix blocks $x^{(<b)}$ to be reused rather than recomputed.
An illustrative example of the difference between full and block diffusion models sampling is given in \cref{fig:block_full}.
This factorization anchors indices and conventions at block boundaries, reduces length sensitivity, and retains parallel token updates within each block while enabling variable-length generation via block-level stopping.

Previous unimodal MDMs~\cite{fastdllm, wu2025fast} use small block sizes (4--32 tokens) to minimize the performance gap with autoregressive models on text-only tasks. By utilizing the properties of the OCR task and equipping the model with bidirectional attention across blocks, we are able to scale the block size from $32$ to $256$ tokens. Finally, this work is the first to apply block-causal masking both during training and inference in the multimodal VLM settings.

\section{Experiments}
\label{sec:experiments}

\subsection{DODO}


We instantiate our proposed framework as \textbf{DODO} (\underline{D}iscrete \underline{O}CR \underline{D}iffusion M\underline{o}dels), built upon the Qwen2.5-VL-3B architecture~\citep{qwen25vl}.
We train DODO on \texttt{olmOCR-mix-1025}~\citep{olmocr}, a large-scale OCR dataset comprising approximately 270K document-text pairs derived from PDFs.
The dataset covers diverse document types, including academic papers, books, reports, and web pages, with text extracted using a combination of PDF parsing and OCR pipelines.
While block-based diffusion has previously been explored for text-only models~\cite{blockdiffusion, wu2025fast}, we are the first to adapt this paradigm to the multimodal domain.

DODO employs a \emph{block-causal} attention mask: tokens within the active block $x^{(b)}$ attend bidirectionally to one another and to all previous blocks $x^{(<b)}$, but attention from previous blocks to the current one is masked.
This strict causality ensures that the representations of the committed prefix remain fixed, enabling the use of an exact Key-Value (KV) cache.
Consequently, only the active block needs to be computed at each step, resulting in significant speedups.
Furthermore, unlike autoregressive models, the diffusion framework also permits full bidirectional attention across blocks, which we show in \cref{sec:standard_vs_block} can further improve accuracy by enabling larger effective block sizes.




\subsection{Implementation Details}
\label{app:implementation}

We fine-tune the model with a maximum sequence length of $8192$ tokens to accommodate dense document texts, retaining the default image preprocessing pipeline.
A visualization of the attention structure is provided in Appendix~\ref{app:attention}.

\paragraph{Training Configuration.}
We train DODO for a total of $200,000$ steps on a node of $8\times$NVIDIA A100 (40GB) GPUs, using a global batch size of $8$.
Optimization is performed using AdamW with a peak learning rate of $5\times 10^{-6}$ and a weight decay of $0.01$.
We employ a Warmup-Steady-Decay (WSD) learning rate scheduler, consisting of a linear warmup for $5,000$ steps, a constant steady phase, and a linear cooldown to zero over the final $20,000$ steps.
To ensure training stability and efficiency, we utilize \texttt{bfloat16} precision throughout the process.

\paragraph{Diffusion Setup.}
Following recent advances in discrete diffusion~\citep{wu2025fast, lavida}, we utilize complementary masking during training.
For timestep sampling, we employ stratified uniform scheduling to ensure balanced coverage of the noise levels during training.

\subsection{Evaluation Setup}

\paragraph{Benchmarks.}
We evaluate our models on two distinct benchmarks.
First, we use \texttt{OmniDocBench}~\citep{omnidocbench}, a comprehensive testbed for layout-sensitive transcription containing 290 English documents across 9 diverse types (e.g., academic papers, financial reports), annotated with structured ground truth for text, tables, and formulas.
Second, we evaluate on \texttt{Fox-Page-EN}~\citep{fox}, a dataset of 112 document pages focused exclusively on pure text without figures or tables.
This combination allows us to assess performance on both complex, multimodal document layouts and standard dense text transcription.

\paragraph{Metrics.}
We assess performance using two metrics. First, we report accuracy using the Normalized Edit Distance (NED) between the predicted and ground-truth text, with lower scores indicating higher fidelity. Second, to quantify inference efficiency, we measure throughput as the number of generated Tokens Per Second (TPS) for each model.


\paragraph{Baselines.}
We compare DODO against three model categories:
(1) specialized OCR models, including dots.ocr, DeepSeek-OCR, MinerU 2.0 VLM, MonkeyOCR, MistralOCR, olmOCR, Nanonets-OCR-s, and SmolDocling~\cite{li2025dotsocrmultilingualdocumentlayout,wei2025deepseek,wang2024mineru,monkeyocr,mistral2025ocr,poznanski2025olmocr,nanonets,smoldocling};
(2) general-purpose autoregressive VLMs, specifically the Qwen2.5-VL family~\citep{qwen25vl}, which serves as our backbone; and
(3) diffusion VLMs, represented by Dimple, LaViDa, and LLaDA-V~\cite{dimple,lavida,lladav}.







\subsection{Main Results}
\label{sec:main_results}

\begin{table}[t]
\centering
\caption{
\textbf{DODO results.}
OCR performance on the English subset of OmniDocBench and Fox-Pages.
Bold numbers indicate the best value across each model type (MDM and AR).
}
\label{tab:main}
\small
\begin{tabular}{@{}lccc@{}}
\toprule
 & & \multicolumn{2}{c}{Edit Distance $\downarrow$} \\
\cmidrule(lr){3-4}
Method & Size & OmniDoc & Fox \\
\midrule
\multicolumn{4}{l}{\textit{Specialized OCR}} \\
dots.ocr & 3B & \textbf{0.032} & 0.034 \\
DeepSeek-OCR & 3.4B & 0.049 & 0.100 \\
MinerU 2.0 VLM & 0.9B & 0.045 & - \\
MonkeyOCR-pro & 3B & 0.058 & 0.084 \\
Mistral OCR & - & 0.072 & \textbf{0.013} \\
olmOCR & 7B & 0.097 & 0.023 \\
Nanonets-OCR-s & 3B & 0.134 & - \\
SmolDocling & 256M & 0.262 & 0.022 \\
\midrule
\multicolumn{4}{l}{\textit{Autoregressive VLMs}} \\
Qwen 2.5 VL & 72B & 0.092 & 0.039 \\
Qwen 2.5 VL & 7B & 0.135 & 0.025 \\
Qwen 2.5 VL & 3B & 0.184 & 0.051 \\
\midrule
\multicolumn{4}{l}{\textit{Diffusion VLMs}} \\
Dimple & 7B & 0.856 & 0.932 \\
LaViDa-L & 8B & 0.994 & - \\
LLaDA-V & 7B & 0.524 & 0.336 \\
\textbf{DODO} & 3B & \textbf{0.069} & \textbf{0.038} \\
\bottomrule
\end{tabular}
\end{table}


\cref{tab:main} presents the OCR performance on OmniDocBench (breakdown by document length and type is provided in Appendix~\ref{app:finegrained}) and Fox-Pages.
Compared to prior diffusion-based VLMs, DODO demonstrates a substantial performance leap.
Standard diffusion models such as Dimple, LaViDa, and LLaDA-V struggle significantly with dense document transcription, incurring high error rates ($>0.5$ NED on OmniDocBench).
In contrast, DODO achieves an NED of $0.069$.
While DODO benefits from specialized OCR training, we demonstrate in our ablations (\cref{sec:standard_vs_block}) that data distribution alone does not explain this gap; even when trained on identical OCR data, standard full-sequence diffusion fails to converge on dense text.
This confirms that the improvement is primarily structural: DODO's block-wise constraints effectively mitigate the alignment failures that plague global diffusion models.

Second, DODO proves highly competitive against strong autoregressive and specialized baselines.
On the layout-intensive OmniDocBench, it surpasses its own autoregressive backbone, the Qwen2.5-VL family, across all model scales.
Furthermore, DODO outperforms various specialized models and achieves near-parity with robust engines such as MonkeyOCR \cite{monkeyocr} and Mistral OCR~\cite{mistral2025ocr}.
These results establish that discrete diffusion is a viable, high-performance alternative to the dominant autoregressive paradigm in the challenging domain of dense text recognition.

\subsection{Throughput Analysis}
\label{sec:results_speed}

\begin{figure}
    \centering
    \includegraphics[width=\linewidth]{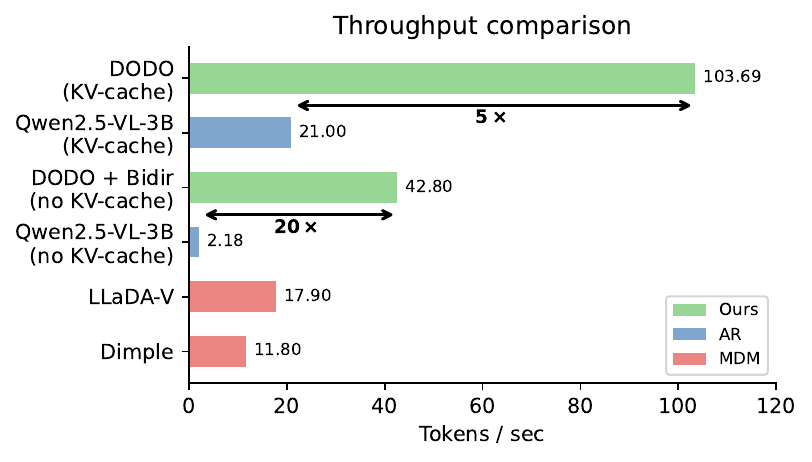}
    \caption{\textbf{Inference throughput comparison.}
DODO leverages parallel decoding, block-causal attention and KV-caching to achieve $\approx 104$ tokens/sec, a $5\times$ speedup over the autoregressive baseline.}
    \label{fig:throughput}
\end{figure}

Figure~\ref{fig:throughput} illustrates the inference throughput across different model architectures.
For qualitative visualizations of the parallel decoding process on dense documents, we refer readers to \cref{fig:token_order} and \cref{app:qualitative}.
Two key trends emerge from this comparison.

First, DODO achieves $\approx 105$ tokens per second, a $5\times$ speedup over the autoregressive Qwen 2.5 VL baseline.
For analysis purposes alone, we isolate the contribution of parallel decoding to the speedup, by running both the AR baseline and DODO+Bidir without KV-caching. The speedup of parallel decoding alone is evaluated as $20\times$ (42.8 vs.\ 2.2 TPS), demonstrating that a significant part of the throughput gain stems from generating multiple tokens per step.
By enabling the use of an exact KV-cache for completed blocks, DODO eliminates the redundant re-computation of prefix representations, maximizing deployment efficiency.
While AR generation requires $L$ sequential steps, DODO decouples latency from sequence length, allowing parallel predictions of multiple tokens to amortize the cost of heavier forward passes.

Second, DODO achieves significantly higher throughput than competing diffusion VLMs.
Prior methods rely on full-sequence attention from the outset, incurring the maximal computational cost proportional to the total length $L$ at every denoising step.
In contrast, DODO combines parallel decoding within blocks with constant cost per block by caching the prefix, allowing it to process generation much faster than standard diffusion baselines, which are immediately burdened by the full sequence complexity. 

\section{Ablation and Empirical Analysis}

We validate the structural design of DODO by isolating the impact of block-based training and analyzing the interaction between block size and caching strategies.
For an ablation of sampling schedules, we refer readers to \cref{sec:sampling}.

\subsection{Vanilla \vs Block Training}
\label{sec:standard_vs_block}

\begin{table}[t]
\caption{
\textbf{Impact of block structure.}
Vanilla MDM fails even with Oracle length; block-wise \textit{training} is essential.
}
\label{tab:ablation_maxlen_blocks}
\centering
\small
\begin{tabular}{@{}lcrrr@{}}
\toprule
\makecell[l]{Train \\ Config} & \makecell{Max \\ Len} & \makecell{Inf. \\ Block} & \makecell{Edit \\ Dist.} & \makecell{TPS} \\
\midrule
Vanilla & Oracle & - & 0.100 & 8.77 \\
Vanilla & 8192 & - & 0.834 & 3.19 \\
Vanilla & 8192 & 32 & 0.951 & 7.47 \\
Block & 8192 & 32 & 0.067 & 29.5 \\
Blk Causal & 8192 & 32 & \textbf{0.069} & \textbf{103.7} \\
\bottomrule
\end{tabular}
\end{table}

\cref{tab:ablation_maxlen_blocks} serves as the empirical validation of the theoretical challenges outlined in \cref{sec:ocr_mdm}. We compare our block-based approach against a ``Vanilla'' baseline: a standard masked diffusion model trained on global sequences (up to 8192 tokens) without block decomposition.

The baseline model exhibits high error rates compared to DODO. Crucially, this performance gap persists even when the model is provided with the \emph{oracle} sequence length. This suggests that the limitation is not solely due to length estimation, but also stems from the positional anchoring inherent to parallel decoding. Attempting to resolve thousands of tokens simultaneously on a fixed canvas creates synchronization risks; because the model cannot adjust the global offset of disjoint text segments, the output becomes fractured.

We further investigate if this issue can be mitigated solely at inference time by applying block decoding to the baseline model.
The results show that restricting the decoding window without a corresponding training objective leads to poor performance.
This contrasts with findings in fast-dLLM~\citep{fastdllm}, where inference-time blocking remained effective for math and coding benchmarks like GSM8K~\citep{gsm8k} and HumanEval~\citep{humaneval}.
We hypothesize that this divergence stems from the nature of the tasks, as unlike OCR, they allow some semantic or syntactic flexibility.
This indicates that block diffusion serves as a necessary structural prior for OCR, conditioning the model to treat the prefix $x^{(<b)}$ as a stable anchor.

Finally, the baseline exhibits significantly lower throughput. This is due to the computational cost of the global canvas: the model must compute attention over the full sequence length (up to 8192 tokens) at every denoising step. In contrast, DODO decomposes the workload, ensuring a more efficient distribution of computational cost.

\subsection{Block Size and Caching Strategies}

\begin{table}[t]
\caption{
\textbf{Block size and caching.}
Approx.\ KV-Cache collapses; block-causal training enables exact caching with $5\times$ speedup.
}
\label{tab:ablation_block_kv}
\centering
\small
\begin{tabular}{@{}lccccc@{}}
\toprule
KV-Cache & Train & Test & Block & Edit Dist. & TPS \\
\midrule
\multicolumn{6}{l}{\textit{No KV-Cache}} \\
 & Bidir & Bidir & 32 & 0.067 & 29.5 \\
 & Bidir & Bidir & 128 & 0.068 & 43.1 \\
 & Bidir & Bidir & 256 & \textbf{0.057} & 42.8 \\
 & Bidir & Bidir & 512 & 0.111 & 35.4 \\
 & Bidir & Bidir & 1024 & 0.192 & 26.5 \\
\midrule
\multicolumn{6}{l}{\textit{Approx. KV-Cache}} \\
 & Bidir & Blk-Causal & 32 & 0.805 & 134.0 \\
 & Bidir & Blk-Causal & 128 & 0.731 & 167.4 \\
 & Bidir & Blk-Causal & 256 & 0.978 & 142.3 \\
\midrule
\multicolumn{6}{l}{\textit{Exact KV-Cache}} \\
 & Blk-Causal & Blk-Causal & 32 & \textbf{0.069} & \textbf{103.7} \\
 & Blk-Causal & Blk-Causal & 128 & 0.089 & 123.1 \\
 & Blk-Causal & Blk-Causal & 256 & 0.177 & 153.4 \\
\bottomrule
\end{tabular}
\end{table}

Table~\ref{tab:ablation_block_kv} investigates the impact of block size and KV-caching strategies on performance. For the standard bidirectional model (No KV-Cache), we observe a non-monotonic trend as the block size increases. Initially, enlarging the block size improves both throughput and accuracy, peaking at an intermediate size ($B=256$). This is because generating more tokens in parallel reduces the total number of sequential inference steps, amortizing the cost of the forward pass. However, further increasing the block size ($B=512, 1024$) leads to diminishing returns in throughput and a notable degradation in accuracy. We attribute this to the recurrence of positional anchoring issues: as the block becomes sufficiently large, it begins to suffer from the same internal synchronization failures that plague the global canvas.

The ``Approx. KV-Cache'' configurations investigate the feasibility of reusing computed keys and values in a bidirectional model without retraining.
Unlike prior findings which observed minimal degradation when freezing history in standard diffusion models~\citep{fastdllm}, our experiments show a sharp accuracy collapse.
We attribute this discrepancy to the rigid nature of the OCR task.

DODO resolves this incompatibility by explicitly training with block-causal masks, enabling exact KV-caching. However, the diffusion framework offers an additional capability unavailable to autoregressive models: bidirectional attention across blocks. Equipping it with bidirectional attention allows the context to dynamically adapt to the current active block by recomputing prefix representations at every step. This reduces the risk of representation drift when generating large blocks, effectively pushing the utilizable block size from $32$ to $256$ tokens and thus further improve accuracy. We observe that while DODO performs best with smaller blocks ($B=32$), equipping it with bidirectional attention uniquely scales to larger blocks ($B=256$), demonstrating that this is a powerful mechanism for enhancing parallel decoding in diffusion models.

\subsection{Inference Efficiency Analysis}
\label{sec:efficiency_analysis}

\begin{figure}
    \centering
    \includegraphics[width=\linewidth]{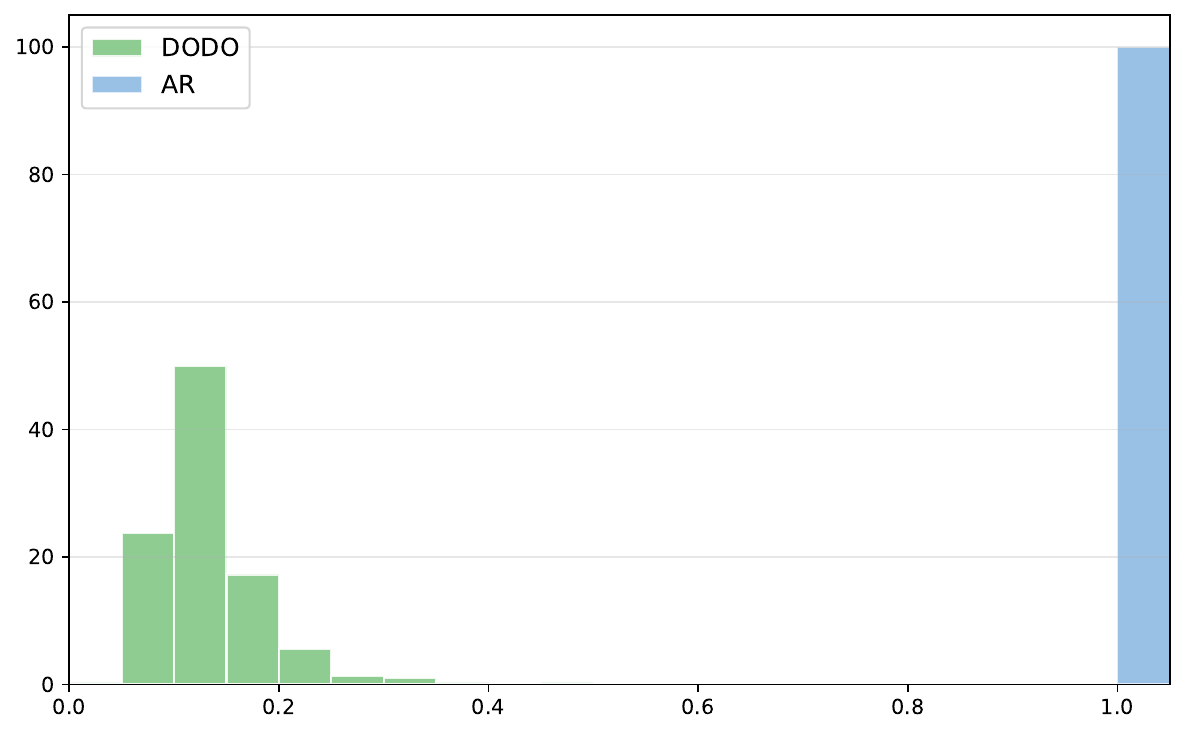}
    \caption{\textbf{Decoding Efficiency.}
DODO requires fewer than $0.1$ steps per token, compressing inference by an order of magnitude vs.\ autoregressive models.}
    \label{fig:steps_per_token}
\end{figure}

To disentangle the source of DODO's throughput advantage, we analyze the number of model forward passes required for generation in \cref{fig:steps_per_token}.
Autoregressive models are structurally bound to a $1:1$ ratio (one step per token).
In contrast, DODO leverages parallel decoding to compress the inference process, typically requiring fewer than $0.1$ steps per token.
This order-of-magnitude reduction in the number of sequential steps explains the throughput results shown in \cref{fig:throughput}.
Although DODO's bidirectional forward pass (without caching) is computationally heavier than a cached autoregressive step, the sheer reduction in the number of calls, generating 10 to 20 tokens per AR step, overcomes the per-step cost.
This allows DODO to outperform optimized autoregressive baselines even without KV caching.

\section{Conclusion}
\label{sec:conclusion}

In this work, we introduce DODO, a framework that unlocks the potential of masked diffusion models to accelerate OCR via parallel decoding.
Our analysis reveals that standard parallel decoding is fundamentally limited by the brittleness of the monolithic canvas, which leads to catastrophic synchronization failures due to length mismatches and positional anchoring.
By decomposing this task into semi-autoregressive blocks, DODO provides a structural remedy that reconciles the stability of causal anchoring with the efficiency of parallel generation.
Empirically, DODO sets a new standard for non-autoregressive OCR VLMs.
It surpasses prior diffusion-based VLMs by an order of magnitude and achieves performance competitive with state-of-the-art specialized and autoregressive systems.
Through block-causal attention and exact KV-caching, DODO achieves a $5\times$ inference speedup over the autoregressive baseline, establishing discrete diffusion not just as a theoretical capability, but as a practical, high-performance alternative for latency-critical applications.
Furthermore, we show that the diffusion framework uniquely enables bidirectional attention across blocks, which further improves accuracy by pushing the utilizable block size from $32$ to $256$ tokens.

\paragraph{Limitations.}
While DODO significantly accelerates inference, its reliance on exact KV-caching enforces a static history, limiting accuracy at larger block sizes compared to the bidirectional variant.
Future work will focus on bridging this gap and exploring diffusion samplers tailored to OCR.

\newpage
\appendix
\onecolumn
\setcounter{figure}{0}
\renewcommand{\thefigure}{\Alph{section}\arabic{figure}}
\section{Attention Structure Visualization}
\label{app:attention}

\begin{figure}[h]
    \centering
    \includegraphics[width=0.6\linewidth]{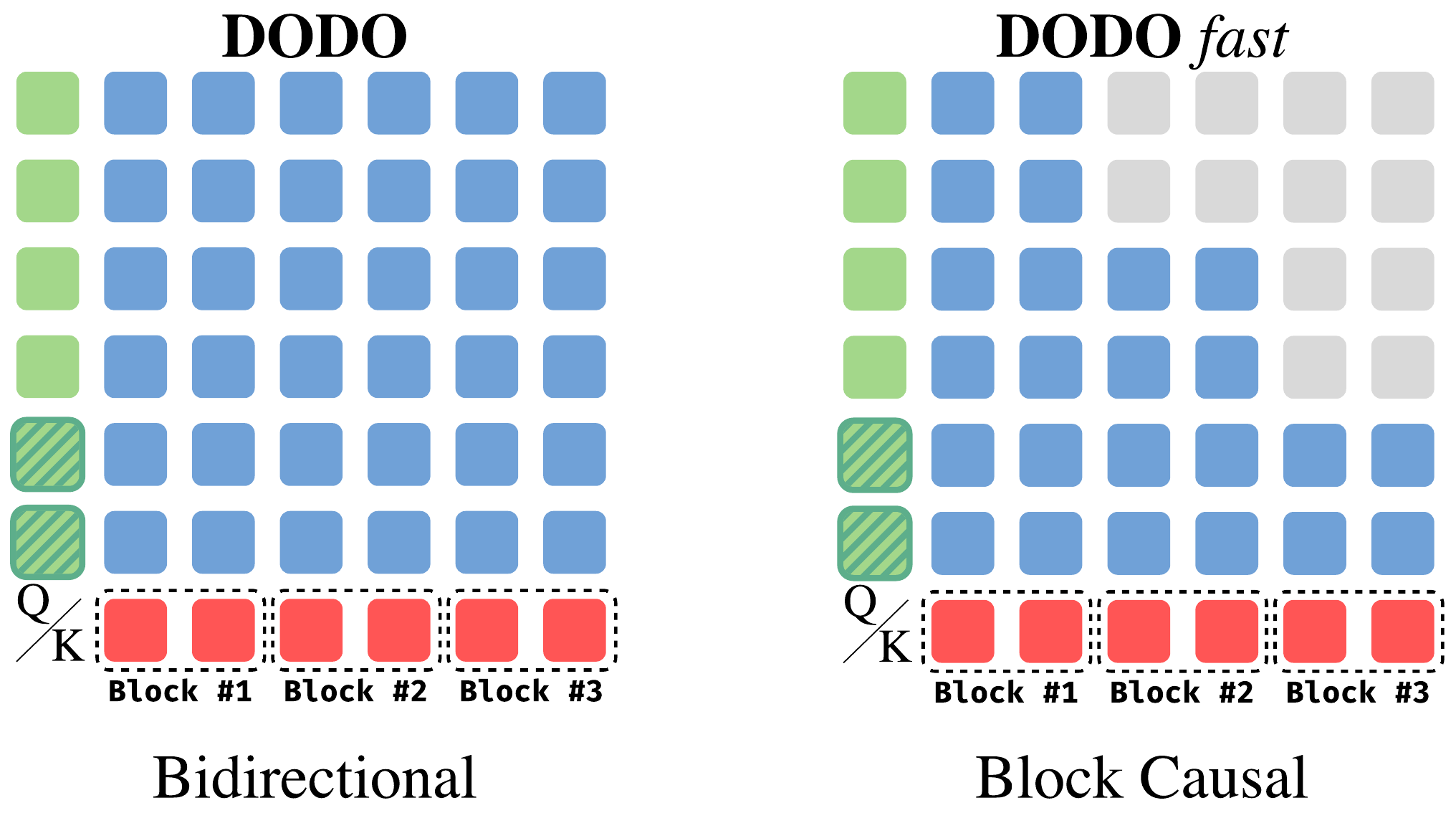}
    \caption{
    \textbf{Visualization of the attention structure.}
    Full bidirectional attention allows prior blocks to attend to the current block (green hatched), meaning their internal representations dynamically adapt during the forward pass.
    Block-causal masking prevents prior blocks from attending to the current block.
    This ensures the history representations remain invariant, enabling exact KV-caching for faster inference.
    }
    \label{fig:attention_masks}
\end{figure}

\section{Additional Ablations}

\subsection{Sampling Strategies}
\label{sec:sampling}

\begin{figure}[t]
\centering
\begin{tikzpicture}
\definecolor{seabornGreen}{HTML}{98D696} 
\definecolor{seabornBlue}{HTML}{80A6CE}  
\definecolor{seabornBack}{HTML}{EAEAF2}  

\begin{axis}[
    width=0.9\linewidth,
    height=8.cm,
    xlabel={Tokens/Second $\rightarrow$},
    ylabel={$\leftarrow$ Edit Distance},
    xmin=0, xmax=320,
    ymin=0, ymax=0.9,
    legend pos=north west,
    legend style={
        font=\small, 
        draw=none,
        fill opacity=0.8,
        text opacity=1,
        fill=seabornBack
    },
    grid=major,
    axis background/.style={fill=seabornBack},
    grid style={white, line width=1pt},
    tick label style={font=\small},
    label style={font=\small},
    axis line style={white},
    tick style={draw=none}
]

\addplot[seabornGreen, line width=1.5pt, mark=*, mark size=2pt] coordinates {
    (137.44, 0.4391)
    (129.10, 0.2846)
    (128.91, 0.2193)
    (126.13, 0.1548)
    (123.40, 0.1446)
    (124.96, 0.1263)
    (120.44, 0.1076)
    (117.59, 0.0785)
    (112.26, 0.0764)
    (111.86, 0.0777)
    (103.69, 0.0686)
};
\addlegendentry{Confidence Thresholding}

\addplot[seabornBlue, line width=1.5pt, mark=square*, mark size=2pt] coordinates {
    (19.08, 0.0578)
    (36.59, 0.0692)
    (67.00, 0.0898)
    (116.45, 0.1805)
    (197.30, 0.5012)
    (301.08, 0.8434)
};
\addlegendentry{Confidence Top-K}

\addplot[only marks, mark=o, mark size=4pt, seabornGreen, line width=1.5pt] coordinates {(103.69, 0.0686)};

\node[font=\tiny, right] at (axis cs:137.44,0.4391) {p=0.5};
\node[font=\tiny, right] at (axis cs:129.10,0.2846) {p=0.6};
\node[font=\tiny, right] at (axis cs:128.91,0.2193) {p=0.7};
\node[font=\tiny, right] at (axis cs:126.13,0.1548) {p=0.75};
\node[font=\tiny, left] at (axis cs:123.40,0.1446) {p=0.8};
\node[font=\tiny, right] at (axis cs:124.96,0.1263) {p=0.85};
\node[font=\tiny, left] at (axis cs:120.44,0.1076) {p=0.9};
\node[font=\tiny, right] at (axis cs:117.59,0.0785) {p=0.95};
\node[font=\tiny, right] at (axis cs:112.26,0.0764) {p=0.97};
\node[font=\tiny, left] at (axis cs:111.86,0.0777) {p=0.98};
\node[font=\tiny, below] at (axis cs:103.69,0.0686) {\textbf{p=0.99}};

\node[font=\tiny, left] at (axis cs:19.08,0.0578) {K=1};
\node[font=\tiny, left] at (axis cs:36.59,0.0692) {K=2};
\node[font=\tiny, left] at (axis cs:67.00,0.0898) {K=4};
\node[font=\tiny, above] at (axis cs:116.45,0.1805) {K=8};
\node[font=\tiny, above] at (axis cs:197.30,0.5012) {K=16};
\node[font=\tiny, above] at (axis cs:301.08,0.8434) {K=32};

\end{axis}
\end{tikzpicture}
\caption{Edit Distance vs Speed for different sampling strategies. \textbf{Confidence Thresholding (Green)} achieves the optimal balance at $p=0.99$, maintaining high accuracy while offering adaptive speedups.}
\label{fig:ed_vs_speed}
\end{figure}

To optimize the inference process, we investigate the impact of the decoding schedule. Given the intolerance of OCR tasks to transcription errors, our primary objective is not merely to maximize speed, but to identify the fastest strategy that maintains high fidelity. We compares two distinct sampling strategies:

\begin{itemize}
    \item \textbf{Confidence Thresholding~\cite{dimple}:} A dynamic strategy that unmasks all tokens whose prediction probability exceeds a fixed threshold $p$. This allows for adaptive step sizes: the model accelerates through clear text and slows down for ambiguous regions. This is the default strategy used in our main experiments (with $p=0.99$).
    \item \textbf{Confidence Top-K~\citep{zheng2023reparameterized}:} A fixed-rate strategy that unmasks exactly $K$ tokens per step, guaranteeing a steady generation pace regardless of model confidence.
\end{itemize}

Figure~\ref{fig:ed_vs_speed} illustrates the performance trade-offs.
In the high-accuracy regime required for OCR, Confidence Thresholding emerges as the superior strategy.
By strictly enforcing a high confidence floor ($p=0.99$), it ensures that the model only commits to tokens when certainty is high.
While other strategies (such as aggressive Top-K) may achieve higher raw throughput, they do so at the cost of unacceptable error rates.
Thus, confidence thresholding provides the optimal balance, maximizing speed strictly within the bounds of usable accuracy.

\section{Fine-Grained Evaluation Breakdown}
\label{app:finegrained}

We provide a detailed breakdown of OmniDocBench results by document length and document type.

\cref{tab:breakdown_length} reveals that DODO's advantage over autoregressive VLMs grows with document length: at 4096+ tokens, DODO achieves $0.079$ vs.\ Qwen 7B's $0.185$, as longer sequences amplify the benefit of parallel decoding. Conversely, for very short documents (0--128 tokens), all methods perform similarly since sequential overhead is minimal.

\cref{tab:breakdown_type} shows that DODO performs consistently across document types, with particularly strong results on PPT slides ($0.014$) and academic papers ($0.050$). The bidirectional variant further improves on structured layouts (Academic, Book) where long-range context aids alignment.

\begin{table}[htb]
\centering
\caption{\textbf{Edit distance by document length (tokens).} Lower is better.}
\label{tab:breakdown_length}
\small
\begin{tabular}{@{}lccccccc@{}}
\toprule
Model & Pages & 0--128 & 128--512 & 512--1024 & 1024--4096 & 4096+ & Overall \\
\midrule
\multicolumn{8}{l}{\textit{Specialized OCR}} \\
MinerU & 275 & 0.328 & 0.166 & 0.027 & 0.032 & 0.030 & 0.041 \\
\midrule
\multicolumn{8}{l}{\textit{Autoregressive VLMs}} \\
Qwen 2.5 VL 7B & 277 & 0.032 & 0.241 & 0.074 & 0.096 & 0.185 & 0.136 \\
Qwen 2.5 VL 3B & 276 & 0.032 & 0.156 & 0.062 & 0.159 & 0.363 & 0.237 \\
\midrule
\multicolumn{8}{l}{\textit{Diffusion VLMs}} \\
Dimple & 269 & 0.343 & 0.697 & 0.750 & 0.868 & 0.900 & 0.857 \\
LaViDa-L & 259 & 0.743 & 0.995 & 0.998 & 1.000 & 1.000 & 0.995 \\
LLaDA-V & 212 & 0.485 & 0.306 & 0.235 & 0.471 & 0.782 & 0.524 \\
\textbf{DODO+Bidir} & 276 & \textbf{0.015} & 0.146 & \textbf{0.025} & \textbf{0.051} & \textbf{0.065} & \textbf{0.057} \\
\textbf{DODO} & 276 & 0.016 & \textbf{0.133} & 0.037 & 0.060 & 0.079 & 0.068 \\
\bottomrule
\end{tabular}
\end{table}

\begin{table}[htb]
\centering
\vspace{-35mm}
\caption{\textbf{Edit distance by document type.} Lower is better.}
\label{tab:breakdown_type}
\small
\begin{tabular}{@{}lcccccccc@{}}
\toprule
Model & Academic & Book & News & Magazine & Textbook & Exam & PPT & Overall \\
\midrule
\multicolumn{9}{l}{\textit{Specialized OCR}} \\
MinerU & 0.015 & 0.049 & 0.047 & 0.043 & 0.052 & 0.045 & 0.261 & 0.041 \\
\midrule
\multicolumn{9}{l}{\textit{Autoregressive VLMs}} \\
Qwen 2.5 VL 7B & 0.098 & 0.189 & 0.306 & 0.119 & 0.067 & 0.115 & 0.020 & 0.136 \\
Qwen 2.5 VL 3B & 0.180 & 0.216 & 0.629 & 0.189 & 0.157 & 0.138 & 0.019 & 0.237 \\
\midrule
\multicolumn{9}{l}{\textit{Diffusion VLMs}} \\
Dimple & 0.900 & 0.872 & 0.880 & 0.836 & 0.805 & 0.871 & 0.404 & 0.857 \\
LaViDa-L & 1.000 & 1.000 & 1.000 & 0.999 & 1.000 & 1.000 & 0.865 & 0.995 \\
LLaDA-V & 0.681 & 0.350 & 0.895 & 0.404 & 0.290 & 0.313 & 0.301 & 0.524 \\
\textbf{DODO+Bidir} & \textbf{0.031} & \textbf{0.048} & 0.109 & \textbf{0.074} & 0.086 & 0.111 & 0.043 & \textbf{0.057} \\
\textbf{DODO} & 0.050 & 0.099 & \textbf{0.093} & 0.088 & \textbf{0.061} & \textbf{0.084} & \textbf{0.014} & 0.068 \\
\bottomrule
\end{tabular}
\end{table}

\section{Document Parsing Examples}
\label{app:qualitative}

\begin{figure*}[t]
    \centering
    
    \QualitativeRow{}{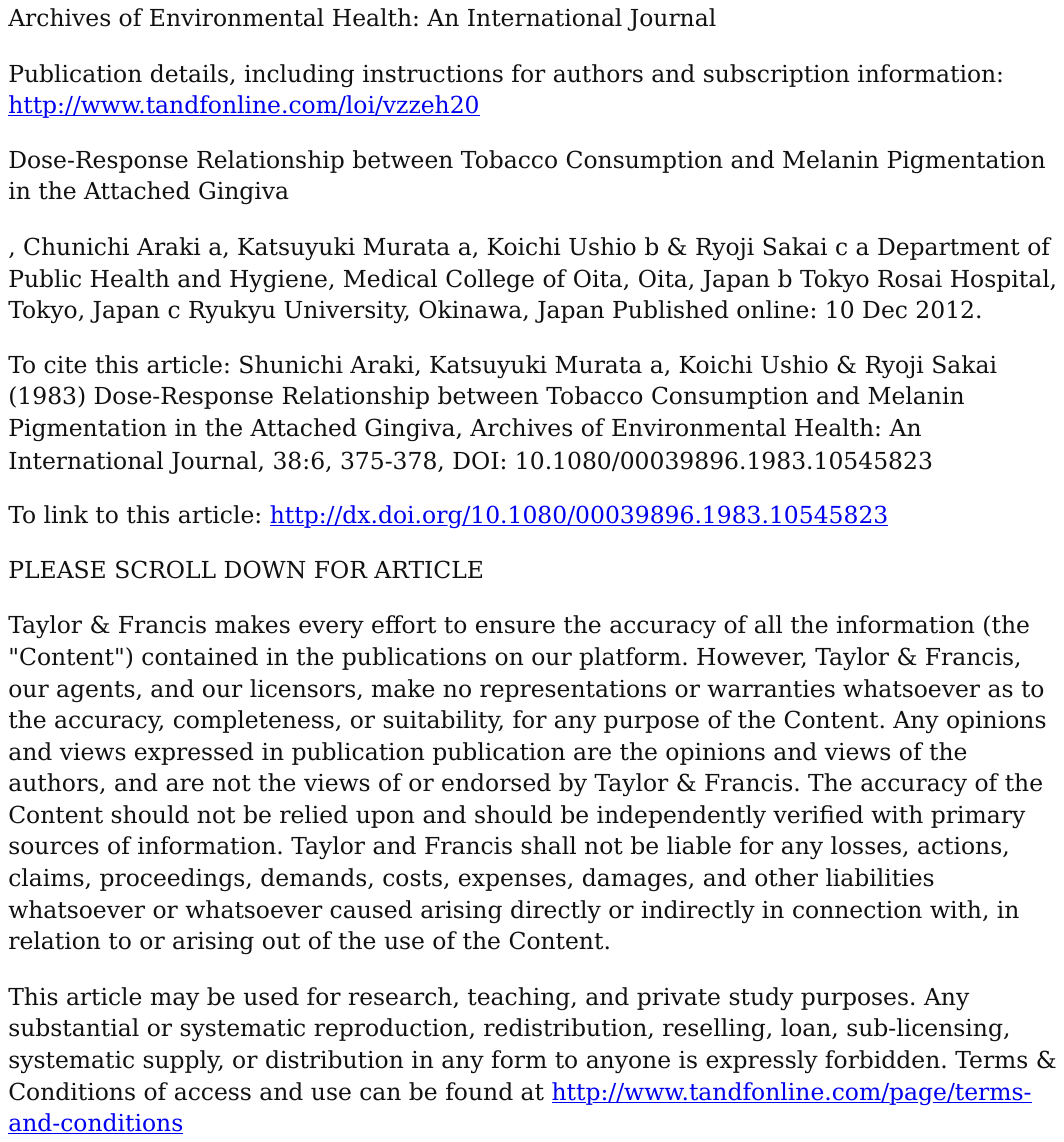}{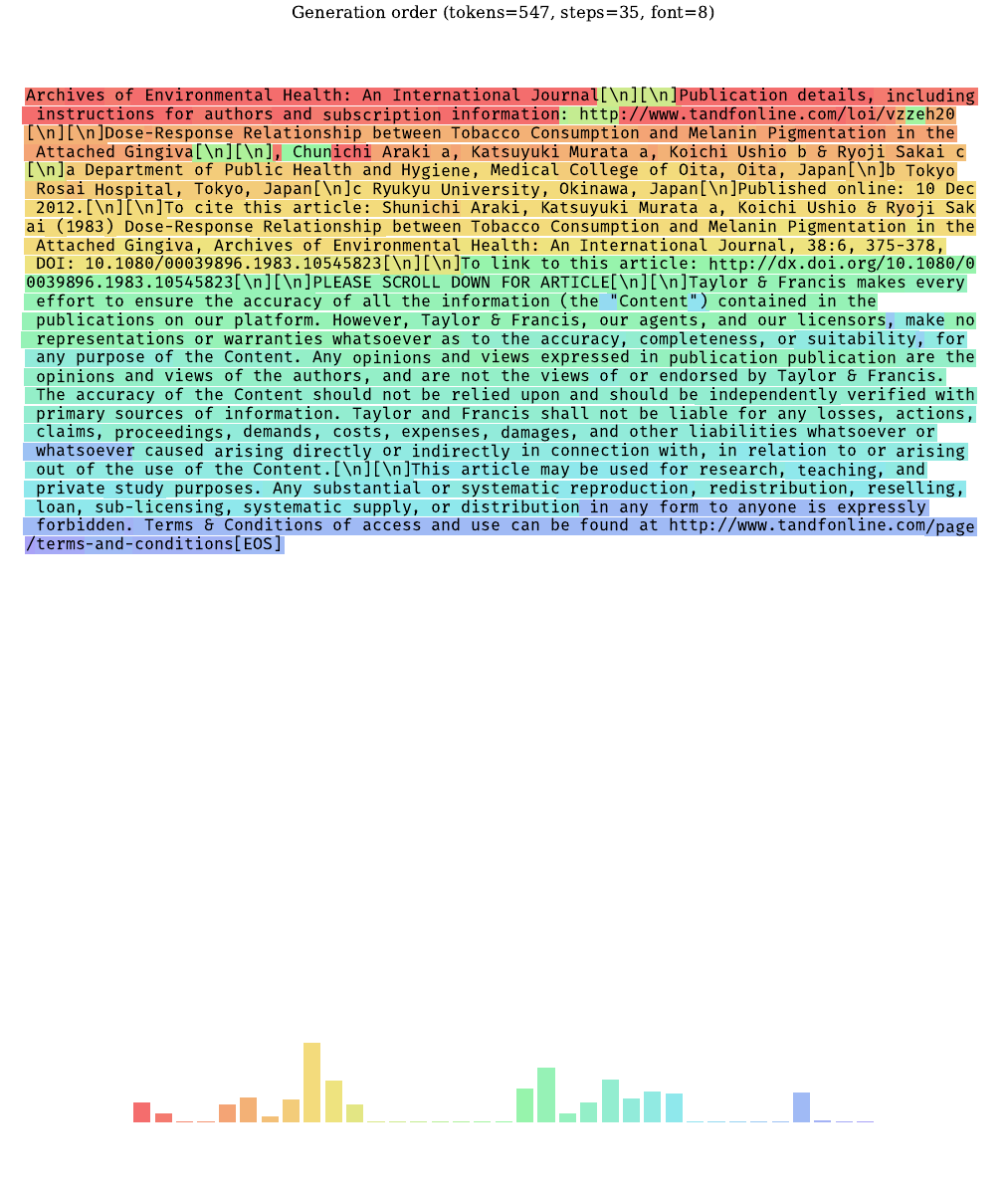}
    
    \QualitativeRow{\detokenize{}}{\detokenize{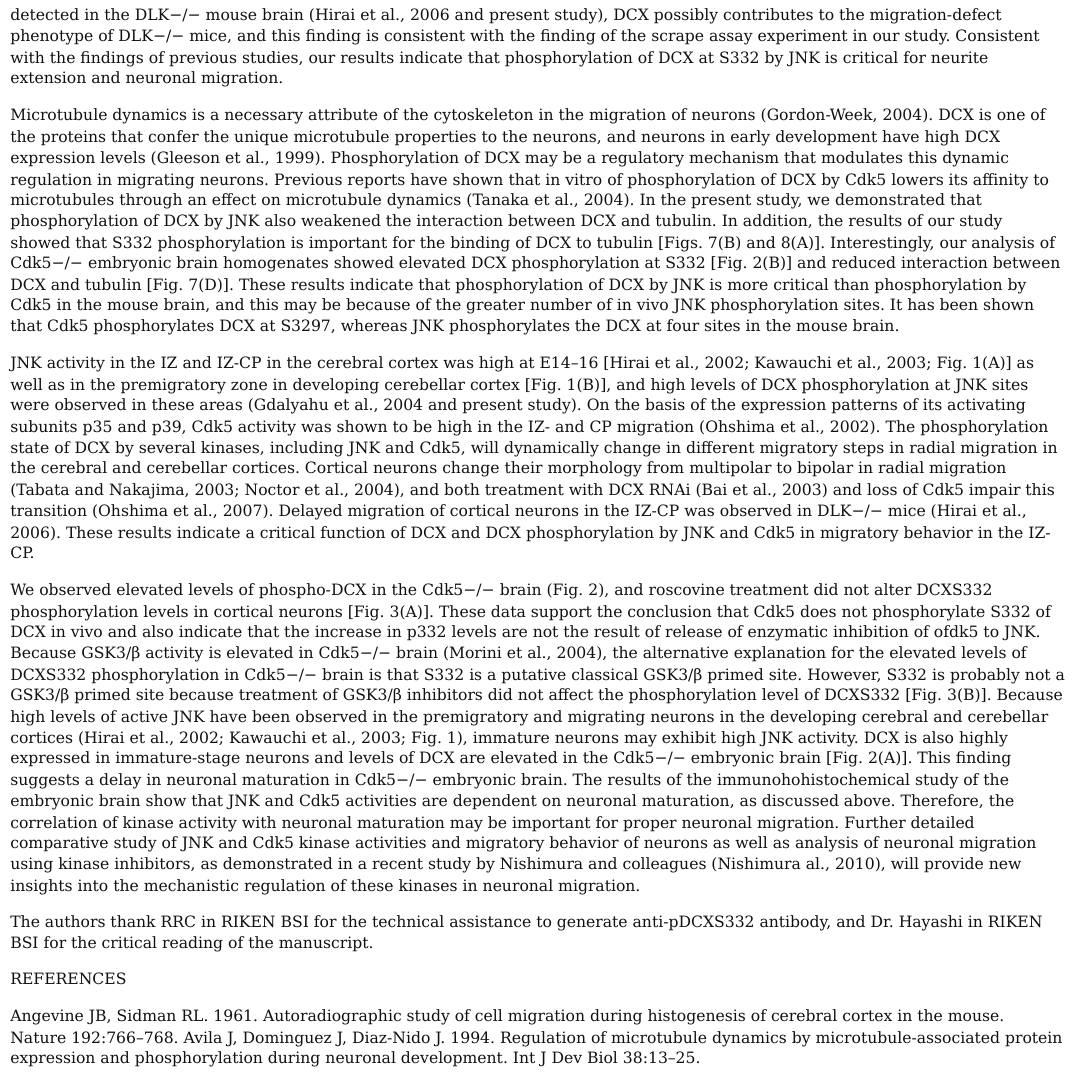}}{\detokenize{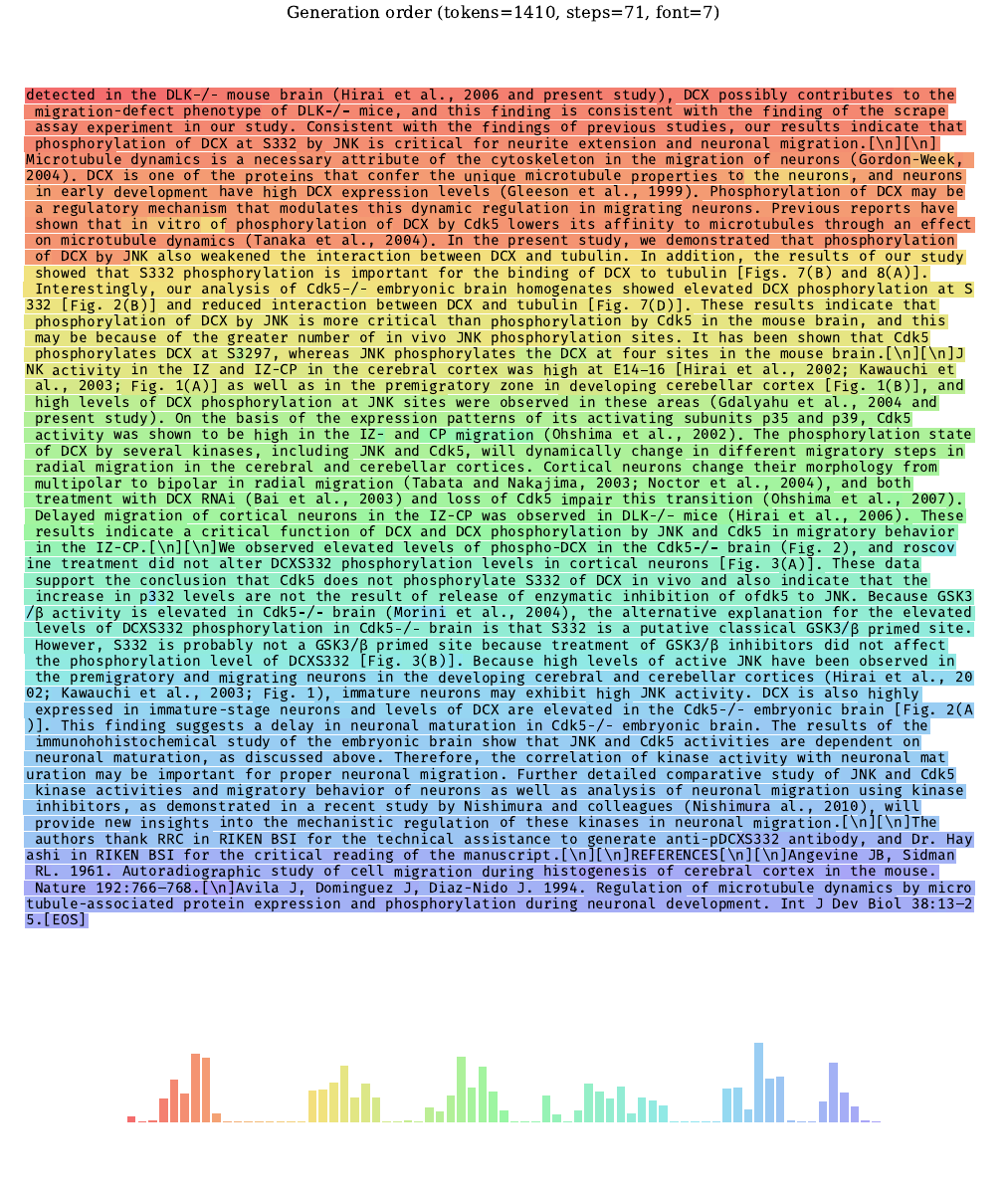}}
    
    \QualitativeRow{\detokenize{}}{\detokenize{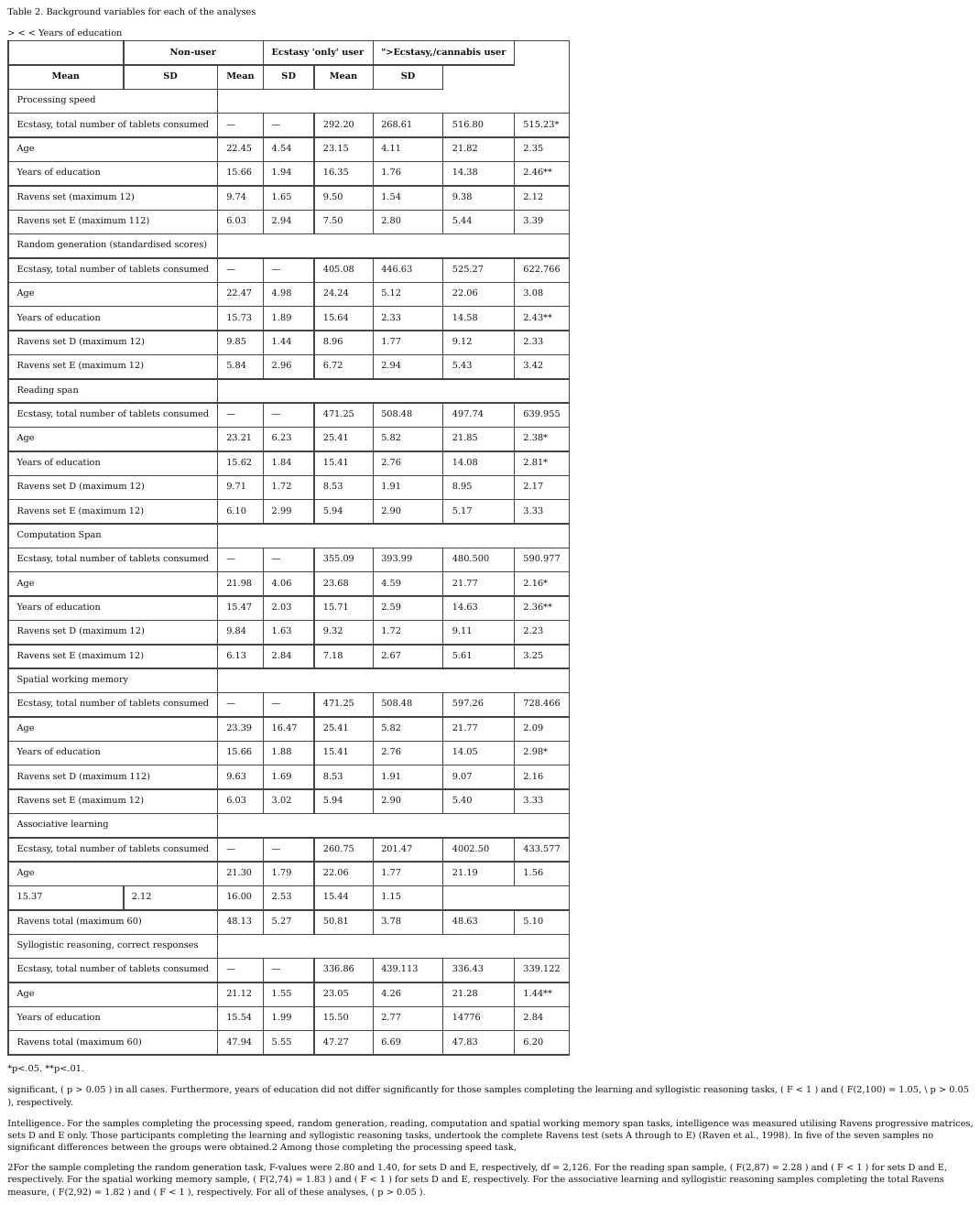}}{\detokenize{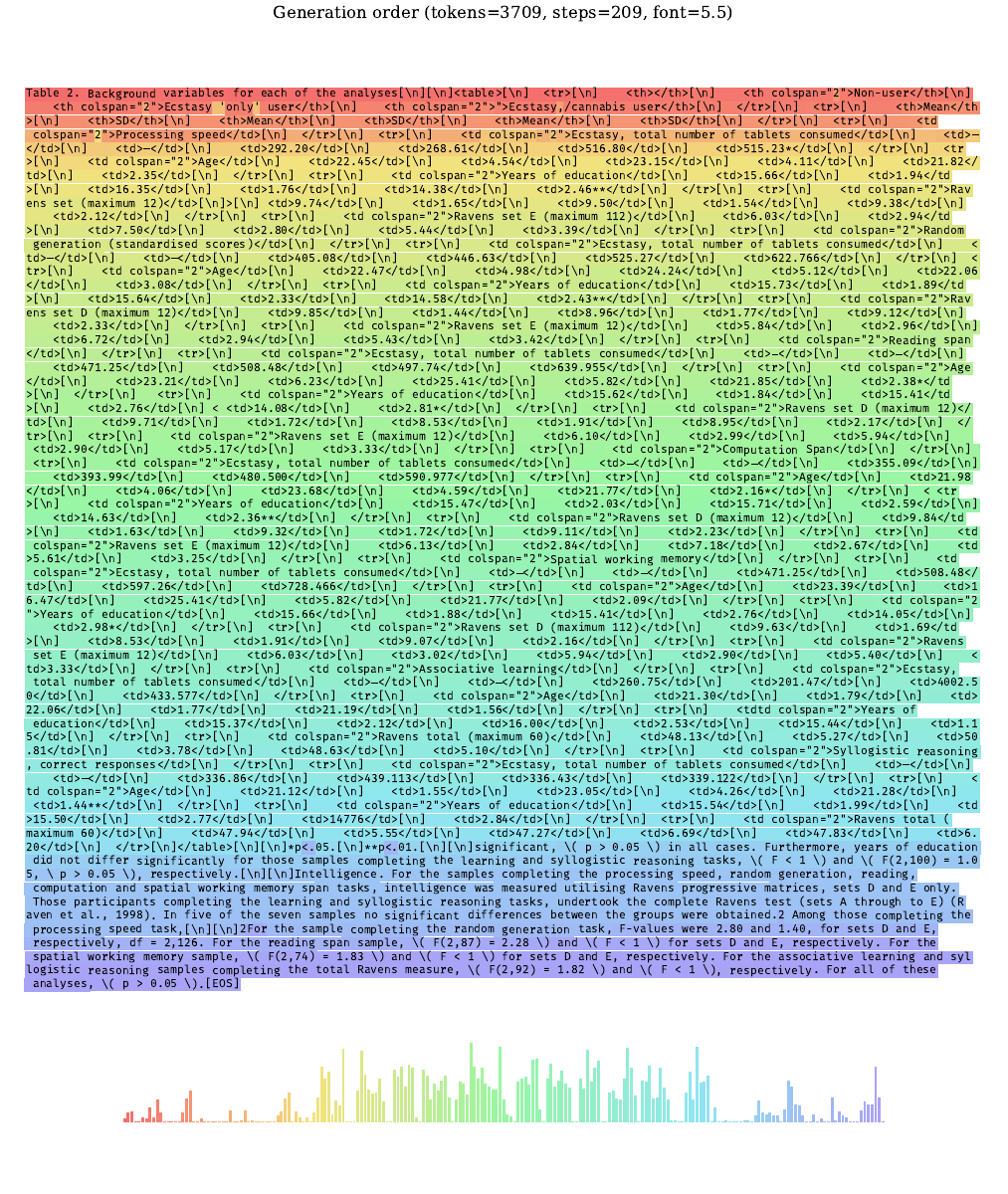}}
    


    \caption{
        \textbf{Qualitative Results.}
        Each row displays a different document from OmniDocBench.
        \textbf{Left:} Input document image.
        \textbf{Center:} DODO's generated transcript rendered to PDF.
        \textbf{Right:} Visualization of the decoding process (heatmap of token commitment order).
        DODO successfully recovers complex layouts, including multi-column text, tables, and mathematical formulas, while maintaining high parallel efficiency.
    }
    \label{fig:qualitative_gallery}
\end{figure*}

\begin{figure*}[t]
    \centering
    
    \QualitativeRow{\detokenize{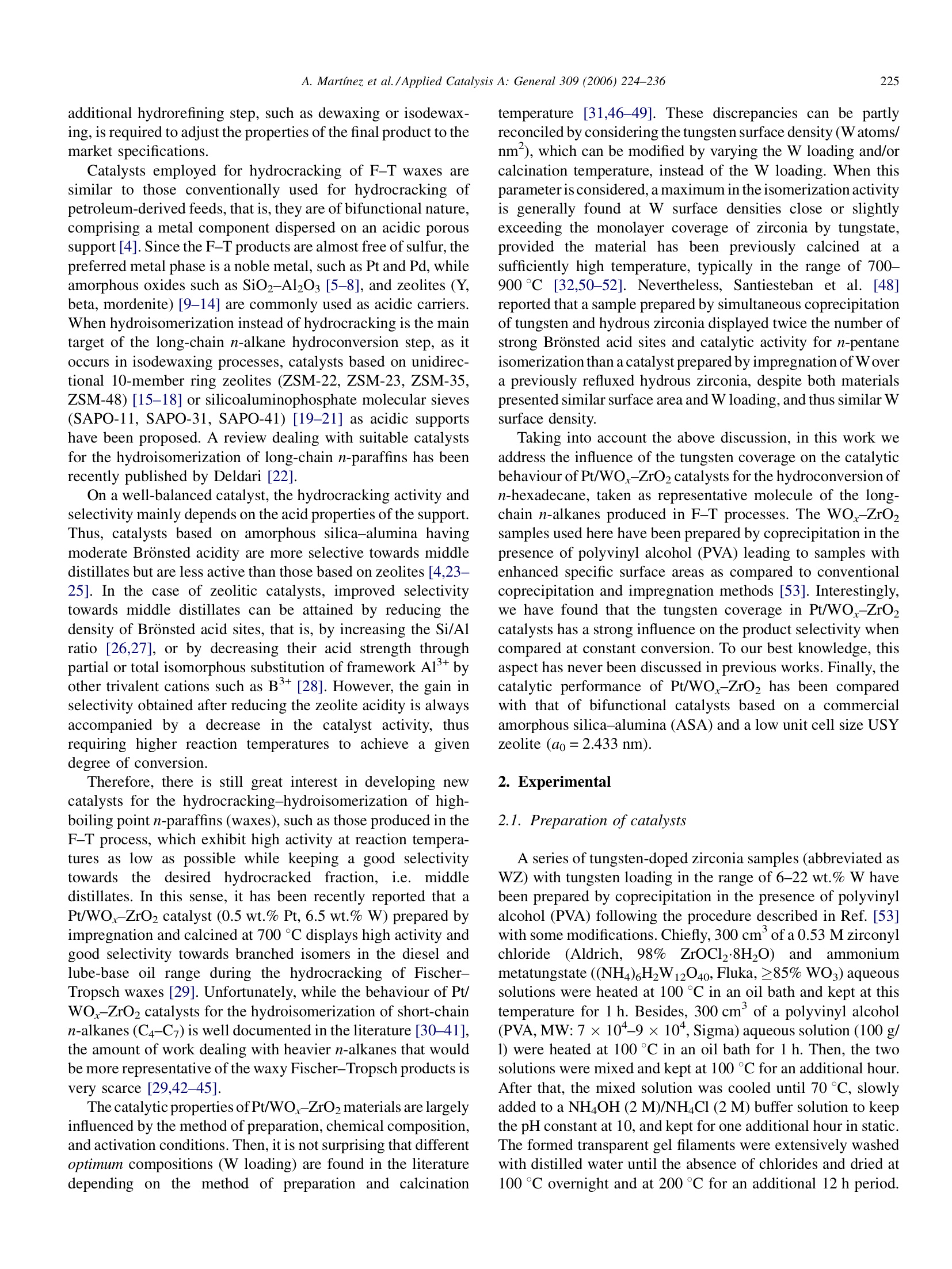}}{\detokenize{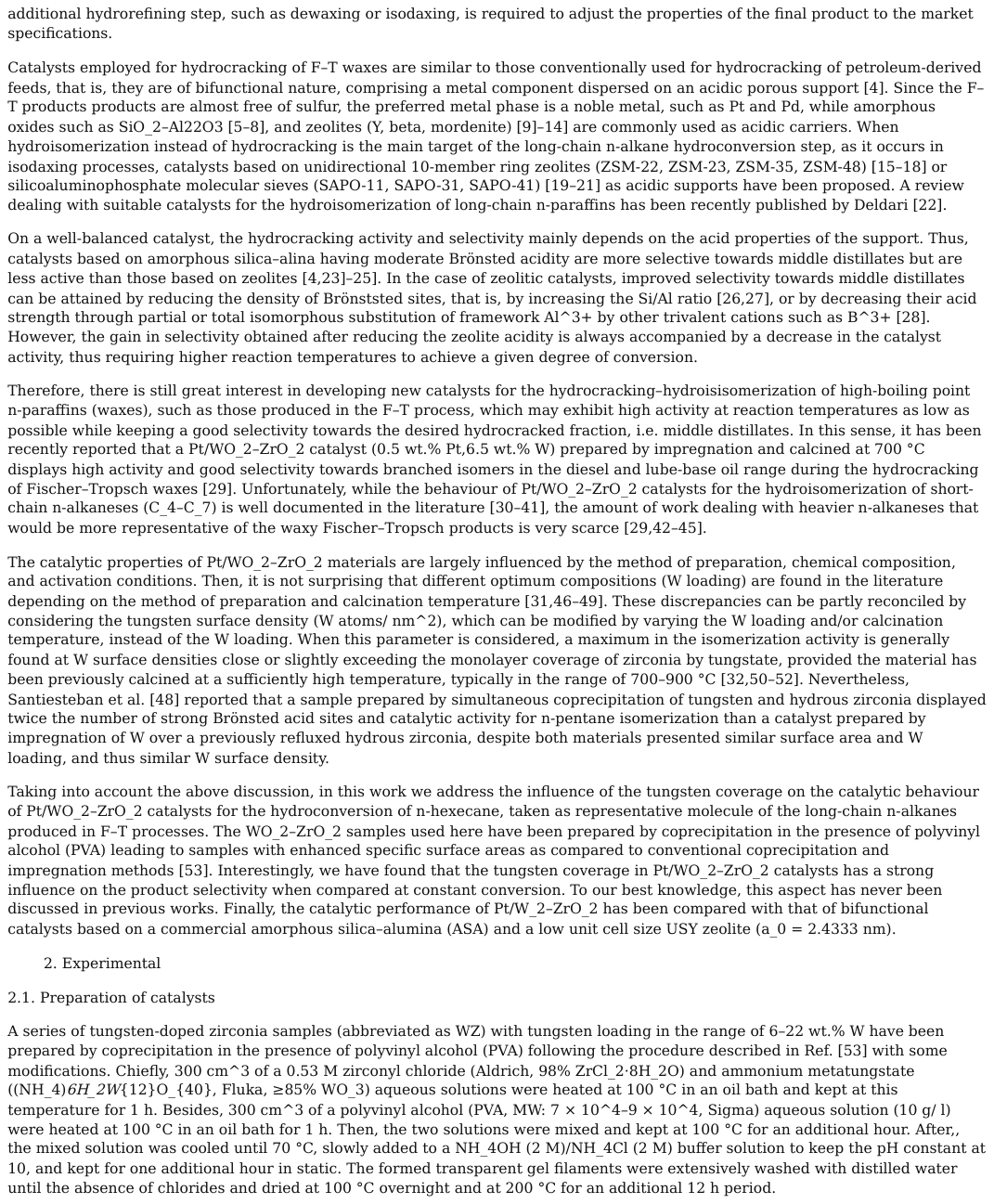}}{\detokenize{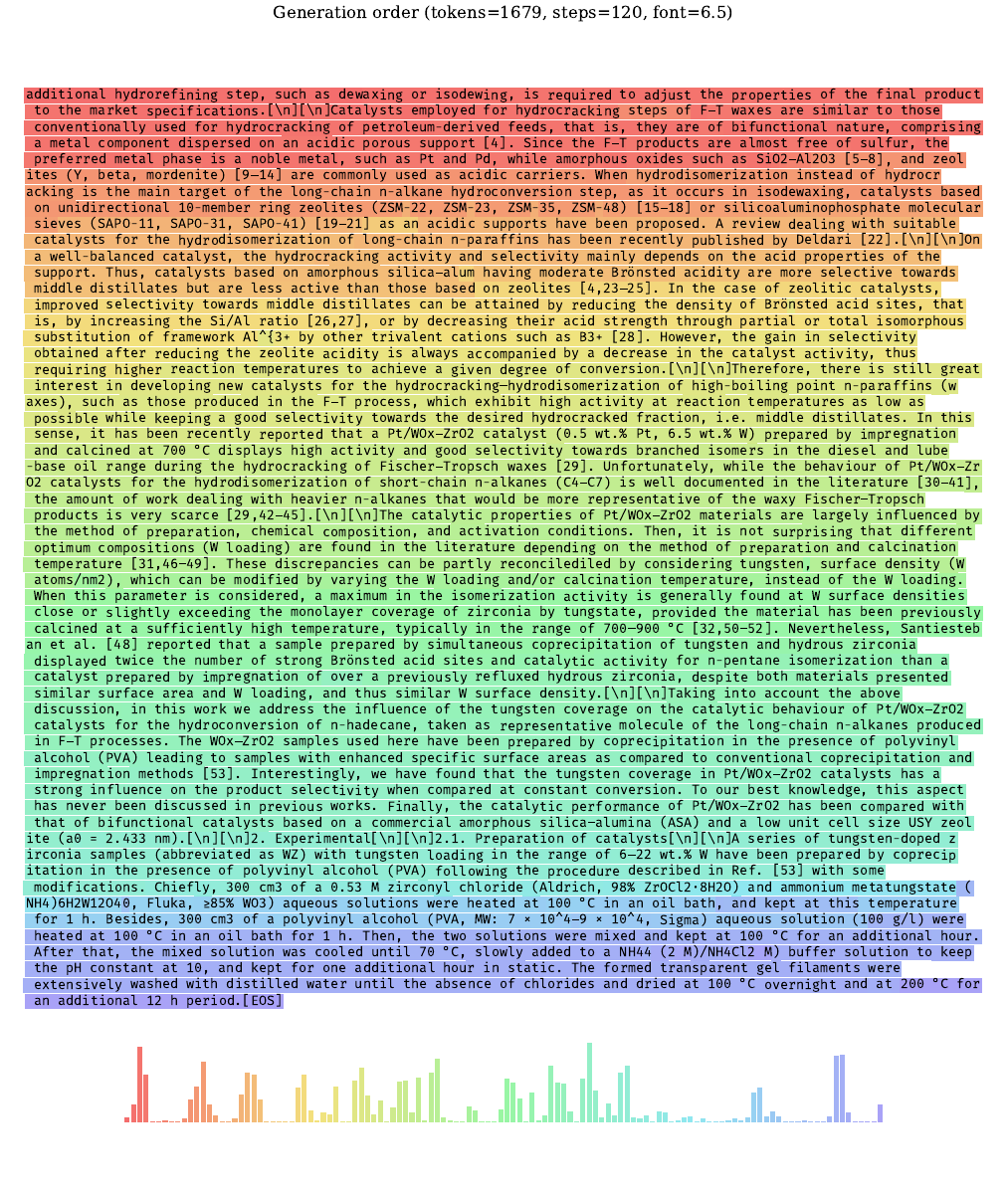}}
    
    \QualitativeRow{\detokenize{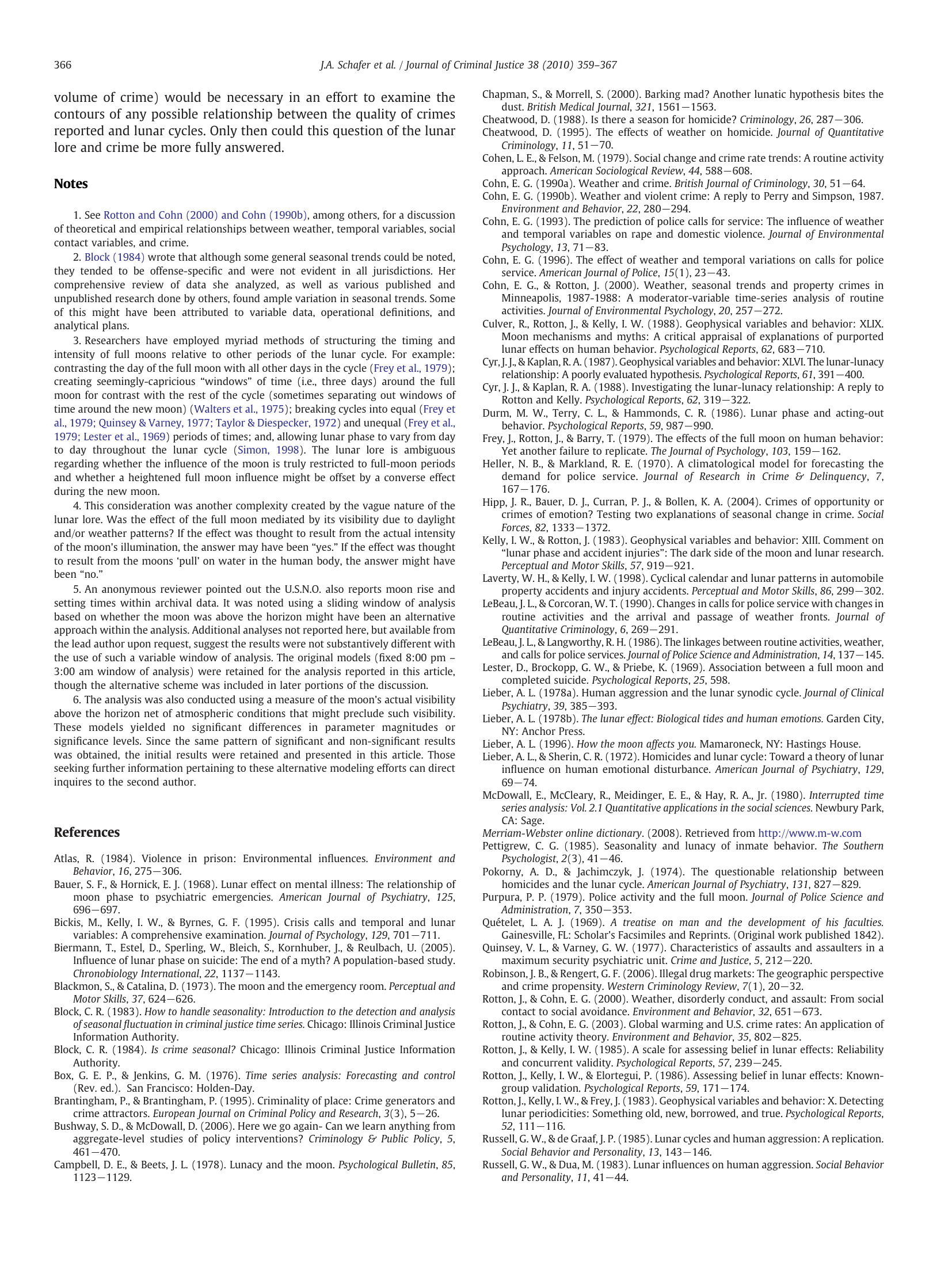}}{\detokenize{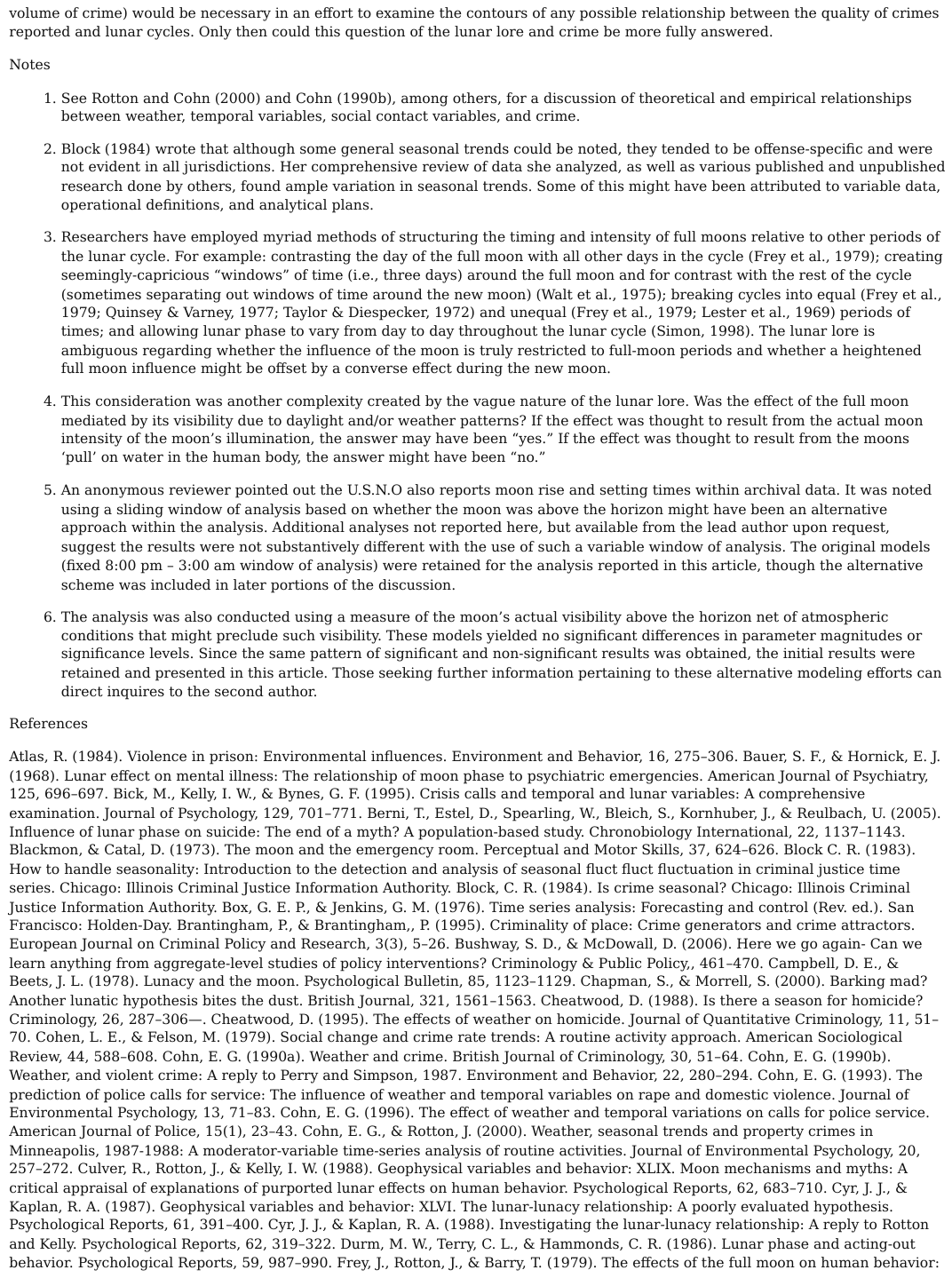}}{\detokenize{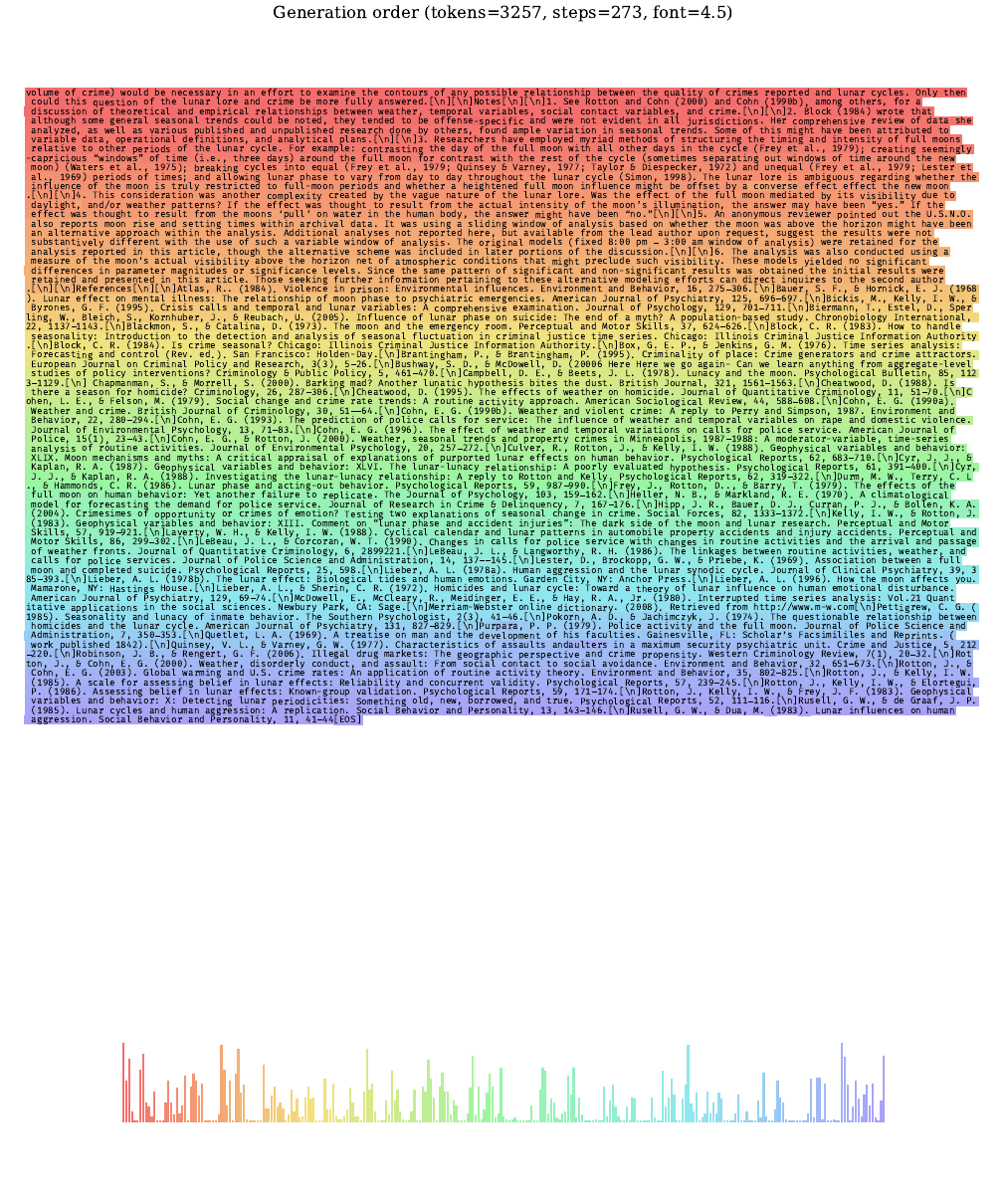}}
    
    


    \caption{
        \textbf{Additional Qualitative Results.}
    }
    \label{fig:qualitative_gallery_more}
\end{figure*}

We present qualitative results in \cref{fig:qualitative_gallery,fig:qualitative_gallery_more}.
Selected from the OmniDocBench dataset, these examples demonstrate DODO's capability to transcribe dense text and preserve complex layout structures, while the accompanying heatmaps visualize the underlying parallel decoding process.

\end{document}